%% file: neurips_2026.tex
\documentclass{article}
\PassOptionsToPackage{numbers,compress}{natbib}

\usepackage[main,final]{neurips_2026}

\usepackage[utf8]{inputenc} 
\usepackage[T1]{fontenc}    
\usepackage[hidelinks]{hyperref}       
\usepackage{url}            
\usepackage{booktabs}       
\usepackage{amsfonts}       
\usepackage{nicefrac}       
\usepackage{microtype}      
\usepackage[table,svgnames]{xcolor}  
\usepackage{amsmath}        
\usepackage{amssymb}        
\usepackage{graphicx}       
\usepackage{wrapfig}        
\usepackage{listings}       
\usepackage[most]{tcolorbox}

\newtcblisting{promptbox}[1][]{
	listing only,
	breakable,
	listing options={
		breaklines=true,
		breakindent=0pt,
		basicstyle=\ttfamily\footnotesize
	},
	colback=gray!10,
	colframe=blue!60!black,
	colbacktitle=blue!60!black,
	boxrule=1pt,
	arc=3pt,
	left=2pt,right=2pt,top=1pt,bottom=1pt,
	fonttitle=\bfseries,
	title=Prompt,
	title style={
		boxsep=4pt,
		top=8pt,
		bottom=8pt,
	},
	#1
}
\graphicspath{{figures/}}

\title{RL-Guided Temporal Localization for Dual-Channel Retrieval in Long-Horizon Agent Memory}

\author{%
	Dabin Sheng\textsuperscript{1},
	Zhe Wu\textsuperscript{1},
	Jinming Zhao\textsuperscript{2},
	Junliang Xing\textsuperscript{1}\thanks{Corresponding author.},
	Yuanchun Shi\textsuperscript{1} \\
	\textsuperscript{1}Key Laboratory of Pervasive Computing, Ministry of Education, \\
	Department of Computer Science and Technology, Tsinghua University \\
	\textsuperscript{2}Qiyuan Lab \\
	{\small\texttt{\{sdb26,wu-z24\}@mails.tsinghua.edu.cn}} \\
	{\small\texttt{\{jlxing,shiyc\}@tsinghua.edu.cn} \quad
		\texttt{zhaojinming@qiyuanlab.com}}
}

\begin{document}

	\setcounter{footnote}{1}
	\maketitle
	\setcounter{footnote}{0}

	\begin{abstract}
		Time-aware retrieval is crucial for long-horizon LLM agents to effectively leverage ever-growing long-term memory. 
		Existing methods focus on lexical or semantic similarity, leading to target memories being crowded out by temporally mismatched yet semantically similar candidates. Even time-aware approaches inject temporal cues only during memory organization or answer generation, overlooking temporal constraints at the retrieval stage, where candidate selection occurs—the root cause of the displacement.
		Inspired by the temporal contiguity effect in human episodic memory, we propose \textbf{TIDER} (\textbf{T}emporal \textbf{I}nterval-driven \textbf{D}ual-channel \textbf{E}vidence \textbf{R}etrieval), a time-aware agentic retrieval framework.
		Central to TIDER is a reinforcement learning (RL)-trained temporal interval locator that infers the temporal windows in which the evidence lies. RL optimizes window localization as a discrete, non-differentiable retrieval decision, 
		using composite rewards to convert sparse answer supervision into fine-grained retrieval feedback.
		These temporal windows first define the temporal retrieval channel; the retrieved candidates are then merged with results from the semantic channel and reranked by proximity to the windows. This temporal grounding enables TIDER to retrieve temporally aligned evidence that would otherwise be drowned out by the volume of semantically similar yet temporally mismatched memories. 
		With a lightweight 1.7B locator, TIDER achieves state-of-the-art accuracy on both long-horizon memory benchmarks---61.52\% on LifeBench (+8.82\%) and 78.96\% on LoCoMo (+3.66\%), which indicates that temporal cues are crucial for pinpointing target evidence amid a vast pool of semantically similar historical events in long-term daily-life memory.
	\end{abstract}

	\section{Introduction}
	\label{sec:intro}
	
	\begin{figure}[t]
		\centering
		\includegraphics[width=\linewidth]{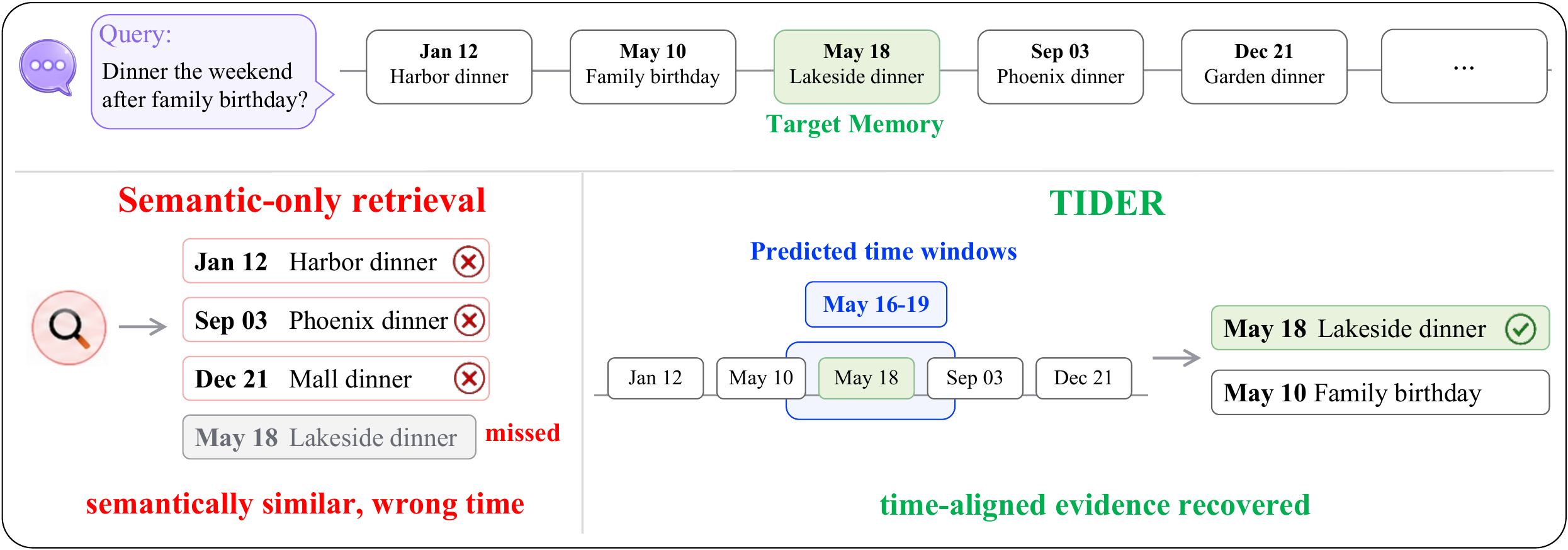}
		\caption{Time-sensitive queries require evidence from a specific window, but semantic-only retrieval
			ranks semantically similar memories from the wrong time. TIDER predicts a temporal window to define the temporal channel, and promotes temporally aligned evidence into the retrieved context.}
		\label{fig:teaser}
	\end{figure}
	
	In recent years, long-term personalized assistants powered by large language model agents have advanced rapidly~\citep{matarazzo2025surveylargelanguagemodels,minaee2025largelanguagemodelssurvey,luo2025largelanguagemodelagent, chen2024personapersonalizationsurveyroleplaying,zhang2025personalizationlargelanguagemodels}. Sustaining such long-horizon interactions requires a foundational capability: retrieving pertinent evidence from an ever-growing history of daily events to address user queries~\citep{chhikara2025mem0buildingproductionreadyai,li2025helloagainllmpoweredpersonalized}. However, daily events are highly homogeneous in surface semantics---\emph{``had dinner with family''} can recur hundreds of times within a year---and the cue that genuinely tells one occurrence from another is often \emph{when} it took place. This makes time-aware retrieval the central bottleneck: the system must locate the right evidence among many semantically similar events recurring at different times. In long-horizon personalized memory benchmarks such as LifeBench~\citep{cheng2026lifebenchbenchmarklonghorizonmultisource} and ATM-Bench~\citep{mei2026accordingmelongtermpersonalized}, 80--90\% of questions involve temporal expressions  such as \emph{``what did I have for lunch after returning from Japan?''}, making time-aware retrieval indispensable rather than incidental.
	
	Existing agentic memory systems incorporate temporal information mostly outside the retrieval stage. Some inject temporal cues during memory organization with timestamps, temporal knowledge graphs, or time-aware structures~\citep{li2026timemtemporalhierarchicalmemoryconsolidation,rasmussen2025zeptemporalknowledgegraph,su2026dialoguetimetemporalsemantic}; others strengthen temporal reasoning during answer generation with neuro-symbolic reasoning, program solvers, or temporal operators~\citep{ge-etal-2025-tremu,10.1145/3726302.3730173,Liang_Zeng_Zhao_Zhao_2026,wu2025chainoftimeline}; recent RL-based methods optimize memory operations or post-retrieval candidate filtering~\citep{yan2026memoryr1enhancinglargelanguage,wang2025memalphalearningmemoryconstruction,du2026memoryt}. None of these methods addresses the root cause: once the correct evidence has been crowded out at retrieval, downstream reasoning and post-retrieval filtering can only operate on an incomplete or contaminated candidate set. Borrowing prior time-aware retrieval is also a poor fit: rule-based normalizers~\citep{zhang-etal-2025-mrag} fail on personal, fuzzy phrasings such as \emph{``the week of her birthday''}, while time-sensitive encoders~\citep{abdallah2025tempretrieverfusionbasedtemporaldense,10.1145/3627673.3679800,qian-etal-2024-timer4} couple temporal and semantic signals in one vector space, demanding extra supervision and degrading semantic recall. Long-term memory, therefore, calls for a retrieval-stage, query-adaptive temporal mechanism decoupled from the semantic encoder.

	Motivated by the temporal contiguity effect in human episodic memory, where retrieval is shaped by temporal distance as well as semantic association~\citep{healey2019contiguity}, we propose \textbf{TIDER} (\textbf{T}emporal \textbf{I}nterval-driven \textbf{D}ual-channel \textbf{E}vidence \textbf{R}etrieval), a time-aware agentic retrieval framework for long-term memory; the basic idea is illustrated in Figure~\ref{fig:teaser}. TIDER decouples temporal cues from the semantic encoder and represents them as temporal windows at the retrieval stage. Central to TIDER is a temporal interval locator $\pi_\theta$ trained via reinforcement learning: given a query and a timeline summary, $\pi_\theta$ decides whether the query is time-sensitive and predicts the windows that contain the target evidence. Window localization is a discrete, non-differentiable decision, so we cast it as an RL problem. The predicted windows define an independent temporal retrieval channel, while the semantic retrieval channel runs in parallel. Results from the two channels are merged and reranked by their distance to the predicted windows, promoting temporally aligned evidence into the top-$k$ and suppressing semantically close but temporally mismatched distractors. $\pi_\theta$ is trained under a composite reward over temporal need, window localization, evidence coverage, and answer correctness. When the query is non-temporal, $\pi_\theta$ closes the temporal channel, and TIDER falls back to the semantic retrieval.

	We evaluate TIDER on LifeBench~\citep{cheng2026lifebenchbenchmarklonghorizonmultisource} and LoCoMo~\citep{maharana-etal-2024-evaluating}, where it consistently outperforms prior agentic memory frameworks, time-aware retrieval methods, and temporal reasoning methods, with the largest gains on hard queries that require distinguishing semantically similar events recurring at different times. Our contributions are:
	\begin{enumerate}
		\item \textbf{A time-aware agentic retrieval framework for long-term memory.} TIDER predicts query-adaptive temporal windows that define a temporal channel parallel to the semantic one, and reranks merged candidates by window distance, resolving the bottleneck that semantically similar events from different times are hard to distinguish at retrieval.
		\item \textbf{A reinforcement-learning algorithm for training the temporal interval locator.} Rewards over temporal need, window localization, evidence coverage, and answer correctness convert sparse answer-only supervision into fine-grained retrieval-stage feedback, enabling the locator to predict evidence-covering windows that improve downstream retrieval.
		\item \textbf{State-of-the-art performance on long-horizon memory benchmarks.} With only a 1.7B locator, TIDER reaches 61.52\% on LifeBench ($+8.82\%$) and 78.96\% on LoCoMo ($+3.66\%$), indicating that temporal cues are crucial for pinpointing target evidence amid a vast pool of semantically similar historical events in long-term daily-life memory.
	\end{enumerate}

	\section{Related Work}
	\label{sec:related}

	\subsection{Long-term Agentic Memory}
	\label{sec:rw-memory}
	
	Long-term agentic memory is an active area of research for LLM agents. One line focuses on the organization of memories: MemGPT~\citep{packer2024memgptllmsoperatingsystems} virtualizes context paging, A-MEM~\citep{xu2026amem} maintains a Zettelkasten-style dynamic index, Mem0~\citep{chhikara2025mem0buildingproductionreadyai} and MemoryBank~\citep{Zhong_Guo_Gao_Ye_Wang_2024} extract and consolidate episodic facts, Zep~\citep{rasmussen2025zeptemporalknowledgegraph} encodes events as a temporal knowledge graph with explicit validity intervals, and TiMem~\citep{li2026timemtemporalhierarchicalmemoryconsolidation} builds a multi-granularity temporal tree. Another line applies reinforcement learning to memory operations: Memory-R1~\citep{yan2026memoryr1enhancinglargelanguage} and Mem-$\alpha$~\citep{wang2025memalphalearningmemoryconstruction} train policies with GRPO~\citep{shao2024deepseekmathpushinglimitsmathematical} to decide when to create or update memory, while Memory-T1~\citep{du2026memoryt} applies RL to multi-session temporal reasoning at the answer stage. All of these methods act either \emph{upstream} of retrieval (deciding what to store and when to update) or \emph{downstream} (filtering candidates after retrieval), and none directly optimizes the retrieval stage itself. TIDER fills this gap by applying reinforcement learning to the generation of retrieval-time temporal constraints rather than to memory operations or post-retrieval reasoning.
	
	\subsection{Time-aware Retrieval and Temporal Reasoning}
	\label{sec:rw-temporal}
	
	Conventional time-aware retrieval models temporal signals inside a vector space: TempRetriever~\citep{abdallah2025tempretrieverfusionbasedtemporaldense} trains separate temporal and semantic encoders, TsContriever~\citep{10.1145/3627673.3679800} trains a temporal query encoder with a temporal-need gate, TSM~\citep{han-etal-2025-temporal} merges multiple time-specifier-conditioned retrievers, and TimeR4~\citep{qian-etal-2024-timer4} trains temporal embedders or rerankers over temporal knowledge graphs. Coupling temporal and semantic signals within a single feature space requires large-scale temporal supervision and degrades semantic recall. Heuristic-based methods such as  MRAG~\citep{zhang-etal-2025-mrag} instead split the query into temporal and semantic parts and rely on hand-crafted templates or date-distance scoring; they cannot parse relative or fuzzy personal-history phrasings such as \emph{``the week of her birthday''}. In contrast, TIDER's temporal windows live outside any vector space: the locator emits explicit temporal windows that constrain an independent temporal channel, leaving the semantic encoder untouched. A complementary line strengthens \emph{downstream} temporal reasoning given retrieved evidence: TReMu~\citep{ge-etal-2025-tremu} translates evidence into Python via neuro-symbolic reasoning, CoTime~\citep{wu2025chainoftimeline} uses SQL-style temporal operators, and NeSTR~\citep{Liang_Zeng_Zhao_Zhao_2026} performs neuro-symbolic abductive reasoning over symbolic temporal relations. These methods take the retrieved set as given and ask how to compute over it, while we ask the upstream question of \emph{whether the retrieved set covers the correct temporal windows at all}.

	\section{Method}
	\label{sec:method}
	
	TIDER is a time-aware agentic retrieval framework for long-term memory QA that decouples temporal constraints from the semantic embedding space and expresses them as explicit temporal windows. The pipeline is built around a reinforcement-learned locator $\pi_\theta$: given a query and a timeline summary, $\pi_\theta$ predicts the windows that contain the required evidence (Section~\ref{sec:temporal-constraint}). The predicted windows define an independent temporal channel that runs in parallel with the always-on semantic channel, and the merged candidates are reranked by their temporal distance to the predicted windows (Section~\ref{sec:dual-channel}). The locator is trained with a composite reward over temporal need, window localization, evidence coverage, and answer correctness (Section~\ref{sec:training}). When the query is non-temporal, $\pi_\theta$ closes the temporal channel, and TIDER falls back to the semantic retrieval.

	\begin{figure}[t] \centering \includegraphics[width=\linewidth]{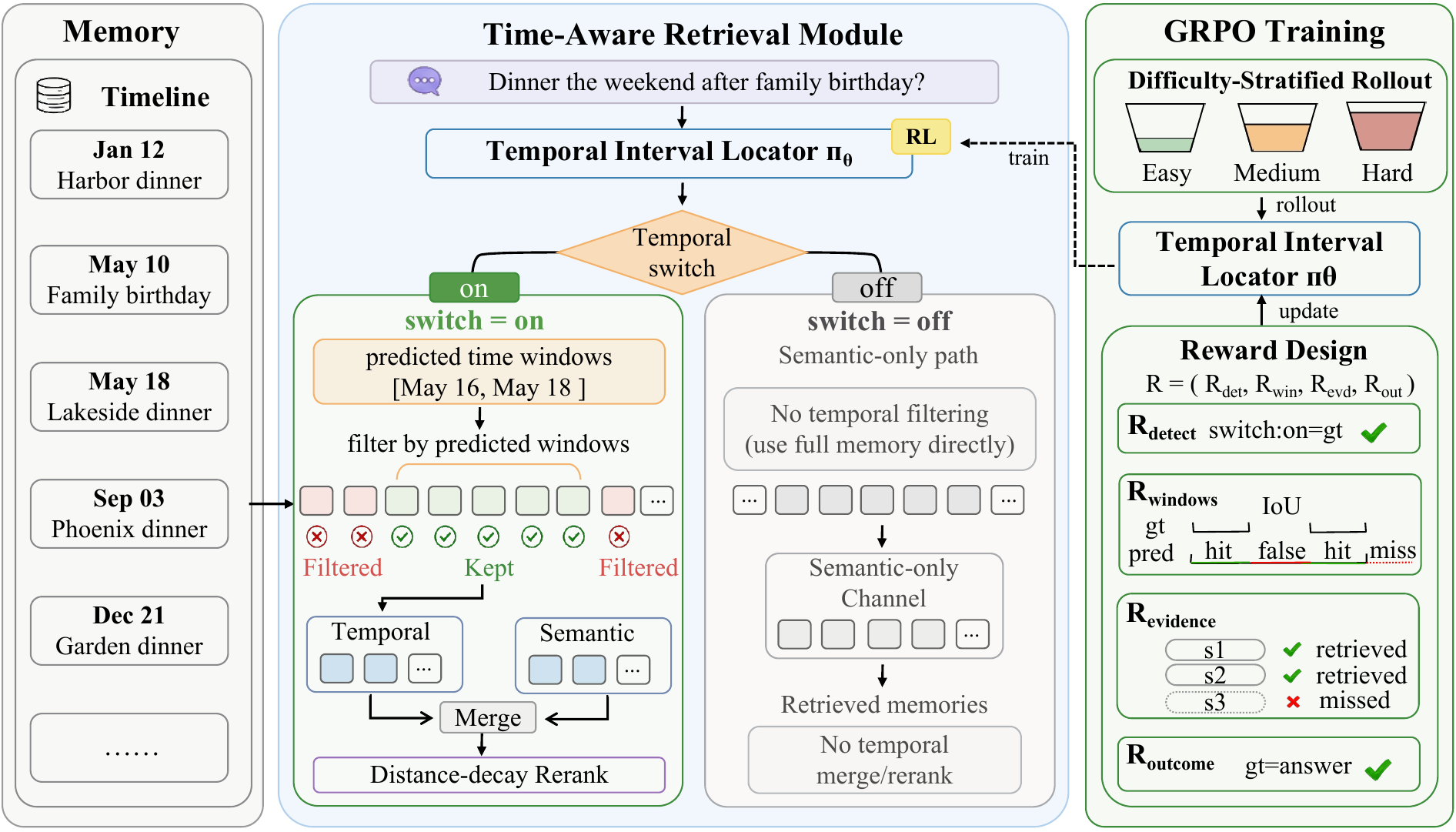} \caption{Overview of TIDER. TIDER uses a temporal interval locator to decide whether a query requires temporal constraints and, if so, to predict explicit temporal windows. The windows define the temporal channel, whose candidates are merged with semantic candidates and reranked by temporal distance; non-temporal queries fall back to the semantic-only path. The locator is trained with GRPO under rewards for switch detection, window localization, evidence coverage, and answer correctness.} \label{fig:framework} \end{figure}
	
	\subsection{Temporal Constraint Generation}
	\label{sec:temporal-constraint}
	
	TIDER takes a user's multi-turn dialogue history $\mathcal{C}$ and a query $q$ as input, and produces an evidence set $\hat{\mathcal{E}}$ for a downstream LLM, which generates an answer $\hat{a}$ conditioned on $\hat{\mathcal{E}}$ and compared against the gold answer $a^*$. This subsection describes the two steps that turn $\mathcal{C}$ into the inputs consumed at the retrieval stage: first, the dialogue is compressed into a temporally anchored event memory pool $\mathcal{M}$, then a locator $\pi_\theta$ rewrites the query's temporal need into an explicit temporal localization $z$.

	We convert the raw dialogue $\mathcal{C}$  into a temporally anchored event memory pool, so that events sharing a temporal neighborhood and semantic similarity are explicitly grouped into the same cluster and become amenable to window-based retrieval:
	\begin{equation}
		\mathcal{M} = \{m_i = (d_i, s_i, \mathcal{I}_i)\}_{i=1}^{N},
		\label{eq:memory-pool}
	\end{equation}
	where $d_i = [d_i^{\text{start}}, d_i^{\text{end}}]$ is cluster $i$'s date span, taken as the union of its members' timestamps; $s_i$ is the cluster's event summary; and $\mathcal{I}_i$ is the set of raw memory items inside the cluster. The construction is performed by an LLM that extracts items with timestamps from each dialogue turn, then clusters them first by date and then by semantic similarity; Appendix~\ref{app:memory-pool} provides the construction details and corpus-level cluster statistics.
	
	Implicit or relative temporal phrasings such as ``the week of her birthday'' or ``after the last get-together'' are hard to cover with rule templates and hurt semantic retrieval when forced into a shared embedding. We therefore introduce a locator $\pi_\theta(\cdot \mid q, \mathcal{S}_q)$ that takes the query $q$ together with a query-adaptive timeline summary $\mathcal{S}_q = \{(d_i, s_i)\}_{i \in \mathcal{N}_q} \subseteq \mathcal{M}$, and emits a structured temporal localization
	\begin{equation}
		z = (\texttt{detect}, W_\text{pred}), \qquad W_\text{pred} = \{w_\ell = [w_\ell^{\text{start}}, w_\ell^{\text{end}}]\}_{\ell=1}^{L},
		\label{eq:control-signal}
	\end{equation}
	where $\texttt{detect} \in \{\text{true}, \text{false}\}$ gates the temporal channel and $W_\text{pred}$ is a set of temporal windows where the evidence may lie. Because the constraint is expressed as explicit windows rather than vectors, TIDER neither needs to train a time-sensitive encoder nor contaminates the semantic embedding; $\pi_\theta$ is trained via GRPO~\citep{shao2024deepseekmathpushinglimitsmathematical} under the composite reward of Section~\ref{sec:training}.

	With $\mathcal{M}$ from preprocessing and $z$ from the locator, TIDER's retrieval can be formalized as a scoring function conditioned on the temporal localization,
	\begin{equation}
		f : (q, \mathcal{M}; z) \mapsto \hat{\mathcal{E}},
		\label{eq:retriever}
	\end{equation}
	whose computation is described in Section~\ref{sec:dual-channel}: it uses $\texttt{detect}$ and $W_\text{pred}$ inside $z$ to drive parallel semantic and temporal recalls, merges their candidates, and reranks the merged set by date distance to $W_\text{pred}$, yielding the final top-$k$ set $\hat{\mathcal{E}}$.

	\subsection{Dual-channel Constrained Retrieval}
	\label{sec:dual-channel}
	
	A single-channel retriever cannot balance semantic and temporal evidence: a semantic-only channel floods the top-$k$ with semantically similar events from wrong times, while a temporal-only channel misses temporally distant yet relevant evidence. TIDER therefore runs the two channels in parallel and merges their candidates. The semantic channel encodes the query $q$ and each memory $m_i$ into vectors $\mathbf{e}_q$ and $\mathbf{e}_{m_i}$ via the dense retriever and uses their cosine similarity as the semantic score,
	\begin{equation}
		\text{sem}(q, m_i) = \frac{\mathbf{e}_q^\top \mathbf{e}_{m_i}}{\|\mathbf{e}_q\|\,\|\mathbf{e}_{m_i}\|},
		\label{eq:sem}
	\end{equation}
	The temporal channel is gated by $\texttt{detect}$: when $\texttt{detect} = \text{false}$, it is closed and retrieval degrades to only semantic; when $\texttt{detect} = \text{true}$, for each $m_i \in \mathcal{M}$ we check whether its temporal window $d_i$ overlaps any window in $W_\text{pred}$ and keep only the overlapping candidates.
	
	The candidates from the two channels are unified by a single ranking score that combines (i) the semantic similarity, (ii) a boost for entering the temporal channel, and (iii) a temporal distance decay that suppresses semantically-close but temporally-mismatched distractors:
	\begin{equation}
		\text{score}(q, m_i) = \gamma(\text{dist}_i) \cdot \bigl[\, \text{sem}(q, m_i) + \lambda_\text{boost} \cdot \mathbf{1}\!\left[m_i \in \text{temporal channel}\right] \,\bigr],
		\label{eq:score}
	\end{equation}
	Here $\text{dist}_i$ denotes the smallest date distance from $m_i$'s event date $d_i$ to $W_\text{pred}$: zero when $d_i$ overlaps any window, and the number of days to the nearest window boundary otherwise. The coefficient $\gamma$ takes a piecewise form with no decay inside a buffer, linear decay outside the buffer:
	\begin{equation}
		\gamma(\text{dist}_i) = \begin{cases}
			1, & \text{dist}_i \le \tau_\text{buf}, \\
			\max\!\bigl(\gamma_\text{min},\; 1 - \rho\,(\text{dist}_i - \tau_\text{buf})\bigr), & \text{dist}_i > \tau_\text{buf}.
		\end{cases}
		\label{eq:gamma}
	\end{equation}
	Here $\tau_\text{buf}$ is the tolerance buffer that absorbs week-level misalignments, $\rho$ is the per-day decay rate, and $\gamma_\text{min}$ is the floor that keeps distant-yet-relevant evidence from vanishing.  When $\texttt{detect} = \text{false}$, the temporal channel is closed, and the decay is bypassed, so $\text{score}(q, m_i)$ reduces to $\text{sem}(q, m_i)$. The system takes the top-$k$ by $\text{score}$ as $\hat{\mathcal{E}}$ and feeds it to the downstream LLM.
	
	\subsection{Reward Design and GRPO Training}
	\label{sec:training}
	
	The pipeline of Sections~\ref{sec:temporal-constraint}--\ref{sec:dual-channel} is driven by $\pi_\theta$, but $z$ is discrete and blocks end-to-end gradient flow from the retriever; we therefore train $\pi_\theta$ with reinforcement learning. Each training sample carries four gold supervisions: $\texttt{gold\_temporal} \in \{\text{true}, \text{false}\}$; the gold window set $W_\text{gold}$, present only when $\texttt{gold\_temporal} = \text{true}$; the gold evidence $\mathcal{E}^* = \{e_k\}_{k=1}^{|\mathcal{E}^*|}$; and the gold answer $a^*$. Appendix~\ref{app:training} details the supervision provenance and split construction.

	Sparse supervision from answer correctness $r_\text{out}$ alone is insufficient for stable learning of \emph{when to trigger the temporal channel} and \emph{which window to fetch from}: a correct answer obtained from a wrong window would still be reinforced by $r_\text{out}$. We therefore decompose the training signal into four complementary rewards that correspond to different aspects of the retrieval decision: temporal need, window localization, evidence coverage, and answer correctness. With fixed reward weights, these rewards are combined as
	\begin{equation}
		R = w_\text{det}\, r_\text{det} + w_\text{win}\, r_\text{win} + w_\text{evd}\, r_\text{evd} + w_\text{out}\, r_\text{out}, \qquad \sum_\bullet w_\bullet = 1,
		\label{eq:reward}
	\end{equation}
	The detection reward $r_\text{det}$ supervises whether the temporal channel is correctly triggered, given by the agreement between $\texttt{detect}$ and $\texttt{gold\_temporal}$:
	\begin{equation}
		r_\text{det} = \begin{cases}
			+1, & \texttt{detect} = \texttt{gold\_temporal}, \\
			-1, & \text{otherwise}.
		\end{cases}
		\label{eq:r-det}
	\end{equation}

	The window-localization reward $r_\text{win}$ measures the geometric alignment of $W_\text{pred}$ with $W_\text{gold}$: for any window pair $(P, G)$ we take the IoU $\varphi(P, G) = |P \cap G| / |P \cup G|$, and aggregate via a symmetric, bidirectional best match. For non-temporal queries or when $W_\text{pred} = \emptyset$, we set $r_\text{win}=0$.
	\begin{equation}
		r_\text{win} = \tfrac{1}{2}\!\left(\tfrac{1}{|W_\text{gold}|}\!\sum_{G \in W_\text{gold}} \max_{P \in W_\text{pred}} \varphi(P, G) + \tfrac{1}{|W_\text{pred}|}\!\sum_{P \in W_\text{pred}} \max_{G \in W_\text{gold}} \varphi(P, G)\right).
		\label{eq:r-win}
	\end{equation}

	The evidence reward $r_\text{evd}$ feeds the retrieval outcome back to the locator: $r_\text{win}$ only scores the geometric overlap between $W_\text{pred}$ and $W_\text{gold}$, while $r_\text{evd}$ checks whether the retriever conditioned on $W_\text{pred}$ actually places $\mathcal{E}^*$ within top-$k$. An independent LLM judge tests, item by item, whether $e_k$ is semantically covered by $\hat{\mathcal{E}}$, and the evidence reward is the coverage rate
	\begin{equation}
		r_\text{evd} = \frac{1}{|\mathcal{E}^*|}\sum_{k=1}^{|\mathcal{E}^*|} \mathbf{1}\bigl[e_k \text{ covered by } \hat{\mathcal{E}}\bigr].
		\label{eq:r-evd}
	\end{equation}

	The outcome reward $r_\text{out}$ is the direct signal of whether the final answer $\hat{a}$ is correct, evaluated per question type:
	\begin{equation}
		r_\text{out} = \begin{cases}
			\mathbf{1}\bigl[\mathrm{ExactMatch}(\hat{a}, a^*)\bigr], & \text{multiple-choice}, \\
			\mathbf{1}\bigl[\mathrm{maj}_{j=1,2,3}\, \mathrm{judge}^{(j)}(\hat{a}, a^*) = 1\bigr], & \text{open-ended},
		\end{cases}
		\label{eq:r-out}
	\end{equation}
	where $\mathrm{ExactMatch}$ checks string equality after normalization, and $\mathrm{judge}^{(j)}$ denotes three independent LLM-judge calls whose majority vote suppresses single-evaluation variance. Appendix~\ref{app:judge-consistency} audits judge consistency across model families.

	A finite annotated $(q, \texttt{detect}, W_\text{pred})$ set covers neither the long tail of natural-language temporal phrasings nor the actual downstream retrieval and reasoning effects, so we adopt a two-stage scheme of SFT cold-start followed by GRPO online optimization: SFT aligns the locator to a fixed JSON schema and seeds an initial window guess, while GRPO uses $R$ as the policy-gradient signal to self-correct over end-to-end rollouts,
	\begin{equation}
		\mathcal{J}(\theta) = \mathbb{E}_{q,\, \{z^{(j)}\} \sim \pi_{\theta_\text{old}}}\!\!\left[\tfrac{1}{G}\sum_{j=1}^{G} \min\!\bigl(\rho^{(j)} A^{(j)},\; \mathrm{clip}\bigl(\rho^{(j)}, 1-\epsilon, 1+\epsilon\bigr)\, A^{(j)}\bigr)\right] - \beta\, \mathbb{D}_\text{KL}\!\bigl(\pi_\theta \,\Vert\, \pi_\text{ref}\bigr),
		\label{eq:grpo}
	\end{equation}
	Rollouts whose output cannot be parsed by the schema are clamped to $R = R_\text{min}$ and skip the downstream judge; rollouts are U-shape difficulty-stratified by $|W_\text{gold}|$.

	\section{Experiments}
	\label{sec:experiments}
	\subsection{Datasets}
	\label{sec:datasets}
	
	\textbf{LifeBench}~\citep{cheng2026lifebenchbenchmarklonghorizonmultisource} is a benchmark for long-term, multi-source, fragmented personal memory that spans both declarative and non-declarative memory. It contains one-year synthetic personal histories for ten users, with on average roughly 13{,}000 entries per user and a total of 2{,}003 questions spanning five categories: Information Extraction (IE), Multi-hop Reasoning (MR), Nondeclarative Memory Reasoning (ND), Temporal and Knowledge Updating (TKU), and Unanswerable (UA); following the LoCoMo evaluation protocol, the Unanswerable category is excluded from evaluation.

	\textbf{LoCoMo}~\citep{maharana-etal-2024-evaluating} focuses on extremely long multi-session dialogues with QA spanning five reasoning categories—single-hop, multi-hop, temporal, open-domain, and adversarial; following prior work, we evaluate on its 1{,}540-question subset excluding the adversarial category. Expanded benchmark descriptions and evaluation protocols are provided in Appendix~\ref{app:datasets}.
	
	\subsection{Experiments Setup}
	\label{sec:setup}
	
	\paragraph{Baselines.}
	We compare against three families of baselines; aside from Oracle, all methods share the same downstream answer model and top-$k$ retrieval budget. \emph{Agentic memory frameworks:} \textbf{MemOS}~\citep{li2025memosoperatingmemoryaugmentedgeneration} is a memory operating system that builds unified mechanisms across parametric, activation, and plaintext memory types; \textbf{TiMem}~\citep{li2026timemtemporalhierarchicalmemoryconsolidation} is a temporal-hierarchical memory framework that consolidates conversations into a temporal memory tree for long-horizon agents; \textbf{Zep}~\citep{rasmussen2025zeptemporalknowledgegraph} is a memory layer service for LLM agents that synthesizes history into a temporally-aware knowledge graph; \textbf{Mem0}~\citep{chhikara2025mem0buildingproductionreadyai} is a memory-centric architecture that extracts, consolidates, and retrieves salient information from ongoing conversations. \emph{Time-aware retrieval:} \textbf{TsContriever}~\citep{10.1145/3627673.3679800} applies supervised contrastive learning with temporally-constrained negatives to fine-tune the query encoder, routing queries through a classifier to handle temporal constraints. \emph{Temporal reasoning:} \textbf{NeSTR}~\citep{Liang_Zeng_Zhao_Zhao_2026} performs neuro-symbolic abductive reasoning over symbolic temporal representations with reflection and consistency checking. We do not include training-based temporal-reasoning methods such as Memory-T1~\citep{du2026memoryt} as direct baselines, since they fine-tune the response model, whereas our evaluation keeps the downstream LLM fixed across methods. We additionally report \textbf{Oracle}, which directly provides the gold-evidence context, as a non-comparable upper bound. 
	
	
	\paragraph{Implementation.}
	The locator $\pi_\theta$ uses \textbf{Qwen3-1.7B}~\citep{yang2025qwen3technicalreport} as its backbone and is trained with GRPO~\citep{shao2024deepseekmathpushinglimitsmathematical} via the veRL framework~\citep{Sheng_2025}, with KL coefficient $\beta=0.1$ and group size $G=8$; the query-adaptive timeline summary $\mathcal{S}_q$ fed to $\pi_\theta$ is restricted to the top-30 most query-relevant timeline entries for computational efficiency. The semantic channel uses the BGE-M3~\citep{chen2024bgem3} dense retriever together with BM25~\citep{robertson2009probabilistic}; the boost coefficient is $\lambda_\text{boost}=0.3$, and the distance-decay function $\gamma$ uses buffer $\tau_\text{buf}=14$, decay rate $\rho=0.05$, and floor $\gamma_\text{min}=0.3$. All methods retrieve top-$k=20$, except Oracle which directly supplies the gold evidence. The downstream answer model and LLM judge are GPT-5-mini on LifeBench and GPT-4o-mini on LoCoMo, fixed across all baselines for fairness. The evaluation is re-executed three times, and the reported metric is the sample mean $\pm$ sample standard deviation across the three runs. The detailed introduction and implementation of baselines can be found in Appendix ~\ref{app:baselines}.
	We evaluate on the user-0 and user-1 split of LifeBench (408 questions) and on the full LoCoMo set; the remaining LifeBench splits (user-2 through user-9) are used to synthesize training samples, so the training and evaluation users on LifeBench are disjoint, while LoCoMo serves as a zero-shot generalization test. 
	
	\subsection{Results}
	\label{sec:results}
	
	\begin{table}[t]
		\centering
		\caption{Performance comparison on the LifeBench benchmark. Categories include Information Extraction, Multi-hop Reasoning, Nondeclarative Memory Reasoning, and Temporal and Knowledge Updating. Best non-Oracle results are \textbf{bolded}; \underline{underline} indicates second best.}
		\label{tab:lifebench}
		\footnotesize
		\setlength{\tabcolsep}{6.5pt}
		\begin{tabular}{l|cccc|cc}
			\toprule
			\textbf{Method} & \textbf{IE} $\uparrow$ & \textbf{MR} $\uparrow$ & \textbf{ND} $\uparrow$ & \textbf{TKU} $\uparrow$ & \textbf{Overall} $\uparrow$ & \textbf{F1} $\uparrow$ \\
			\midrule
			TsContriever~\citep{10.1145/3627673.3679800}  & 31.28$\pm$1.05             & 23.33$\pm$0.00             & 27.43$\pm$0.60             & 23.91$\pm$0.00             & 27.21$\pm$0.42             & 49.36 \\
			MemOS~\citep{li2025memosoperatingmemoryaugmentedgeneration}         & 54.57$\pm$1.43             & 48.89$\pm$0.48             & 42.01$\pm$1.20             & 31.16$\pm$1.26             & 47.30$\pm$0.88             & 45.90 \\
			Zep~\citep{rasmussen2025zeptemporalknowledgegraph}           & 43.61$\pm$0.40             & 39.44$\pm$0.96             & 32.99$\pm$0.60             & 30.43$\pm$2.17             & 38.40$\pm$0.37             & 38.55 \\
			Mem0~\citep{chhikara2025mem0buildingproductionreadyai}          & \underline{57.99}$\pm$1.05 & 48.33$\pm$2.89             & 50.35$\pm$1.59             & \underline{44.20}$\pm$6.64 & 51.80$\pm$1.26             & 51.94 \\
			TiMem~\citep{li2026timemtemporalhierarchicalmemoryconsolidation}         & 17.58$\pm$1.43             & 17.50$\pm$1.44             & 21.53$\pm$3.01             & 10.87$\pm$2.17             & 17.73$\pm$0.57             & 18.89 \\
			NeSTR~\citep{Liang_Zeng_Zhao_Zhao_2026}         & 56.16$\pm$1.37             & \underline{52.22}$\pm$1.27 & \underline{53.12}$\pm$1.80 & 42.03$\pm$3.32             & \underline{52.70}$\pm$0.65 & \underline{61.25} \\
			\midrule
			\rowcolor{gray!20}
			\textbf{TIDER} & \textbf{67.12}$\pm$0.40 & \textbf{60.83}$\pm$2.10 & \textbf{60.42}$\pm$0.60 & \textbf{47.83}$\pm$2.51 & \textbf{61.52}$\pm$0.86 & \textbf{69.23} \\
			\midrule
			\textit{Oracle (upper bound)} & \textit{87.44$\pm$1.05} & \textit{82.78$\pm$0.96} & \textit{81.94$\pm$3.18} & \textit{74.64$\pm$2.51} & \textit{83.33$\pm$0.74} & \textit{76.79} \\
			\bottomrule
		\end{tabular}
	\end{table}

	\paragraph{Main results.}
	On LifeBench (Table~\ref{tab:lifebench}, 408 questions) and LoCoMo (Table~\ref{tab:locomo}, 1{,}540 questions), TIDER simultaneously attains the best overall accuracy: \textbf{61.52}\% on LifeBench (second-best NeSTR 52.70\%, $+8.82$\%) and \textbf{78.96}\% on LoCoMo (second-best TiMem 75.30\%, $+3.66$\%); the locator is only \textbf{1.7B}, far smaller than the shared downstream answer models GPT-5-mini / GPT-4o-mini, confirming that the gain comes from a structured retrieval-time temporal localization rather than from a stronger response model. Compared with agentic-memory frameworks (MemOS, TiMem, Zep, and Mem0), TIDER achieves the best performance on both datasets. On LifeBench, it outperforms the strongest baseline in this group, Mem0 (51.80\%), by 9.72\%; on LoCoMo, it surpasses the strongest baseline in this group, TiMem (75.30\%), by 3.66\%.
	The time-aware retriever TsContriever performs poorly on both datasets, with a substantially larger gap to TIDER on LifeBench than on LoCoMo. This result supports the concern raised in \S\ref{sec:intro}: coupling temporal and semantic signals in a shared representation space can weaken semantic recall, especially in large-scale long-term memory settings such as LifeBench, where relative and indirect temporal expressions are more prevalent and place greater stress on encoder-fused retrieval. As a temporal reasoning baseline, NeSTR processes temporal information at the answer stage rather than during retrieval. It underperforms TIDER by 8.82\% and 5.48\% on LifeBench and LoCoMo, respectively, highlighting the importance of addressing temporal mismatch before evidence enters the downstream context. The effectiveness of TIDER stems from its dual-channel retrieval architecture with explicit window constraints: by moving temporal constraints upstream into the retrieval stage as an explicit window, TIDER suppresses semantically similar but temporally mismatched candidates without degrading semantic recall, offering a structural advantage over both embedding-stage temporal supervision and reasoning-stage post-processing.
	
	Although TIDER achieves the highest overall accuracy on LoCoMo, its Temporal accuracy and token-level F1 are slightly below TiMem's. Our analysis suggests this gap stems from cross-dataset window-localization mismatch on LoCoMo's sparser timeline: predicted windows cover 76.25\% of gold dates, compared with 93.77\% on LifeBench. Using oracle windows raises Temporal accuracy from 71.65\% to 79.13\% and F1 from 51.72\% to 57.83\%, surpassing TiMem on both.
	
	\begin{table}[t]
		\centering
		\caption{Performance comparison on the LoCoMo benchmark. Reported categories include Single-hop, Multi-hop, Temporal Reasoning, and Open-domain Knowledge. Best results are \textbf{bolded}; \underline{underline} indicates second best.}
		\label{tab:locomo}
		\footnotesize
		\setlength{\tabcolsep}{4.2pt}
		\begin{tabular}{l|cccc|cc}
			\toprule
			\textbf{Method} & \textbf{Single-Hop} $\uparrow$ & \textbf{Multi-Hop} $\uparrow$ & \textbf{Temporal} $\uparrow$ & \textbf{Open-Domain} $\uparrow$ & \textbf{Overall} $\uparrow$ & \textbf{F1} $\uparrow$ \\
			\midrule
			TsContriever~\citep{10.1145/3627673.3679800}  & 77.41$\pm$0.45             & 57.92$\pm$0.33             & 62.93$\pm$0.25             & 39.58$\pm$0.85             & 68.47$\pm$0.16             & 48.94 \\
			MemOS~\citep{li2025memosoperatingmemoryaugmentedgeneration}         & 76.07$\pm$0.10             & 56.85$\pm$0.69             & 69.47$\pm$0.25             & 45.14$\pm$0.49             & 69.24$\pm$0.11             & 45.02 \\
			Zep~\citep{rasmussen2025zeptemporalknowledgegraph}           & 50.42$\pm$0.29             & 42.20$\pm$0.77             & 19.11$\pm$0.29             & 38.19$\pm$0.49             & 41.62$\pm$0.21             & 26.88 \\
			Mem0~\citep{chhikara2025mem0buildingproductionreadyai}          & 62.09$\pm$0.42             & 50.14$\pm$0.32             & 59.25$\pm$0.41             & 37.70$\pm$0.41             & 57.79$\pm$0.34             & 42.52 \\
			TiMem~\citep{li2026timemtemporalhierarchicalmemoryconsolidation}         & 81.43$\pm$0.05             & 62.20$\pm$0.82             & \textbf{77.63}$\pm$0.34    & \textbf{52.08}$\pm$0.74 & \underline{75.30}$\pm$0.16 & \textbf{54.40} \\
			NeSTR~\citep{Liang_Zeng_Zhao_Zhao_2026}         & \underline{82.60}$\pm$0.25 & \underline{63.00}$\pm$1.02 & 66.56$\pm$0.18             & 47.57$\pm$0.60             & 73.48$\pm$0.23             & \underline{53.54} \\
			\midrule
			\rowcolor{gray!20}
			\textbf{TIDER} & \textbf{86.44}$\pm$0.79 & \textbf{74.11}$\pm$2.41 & \underline{71.65}$\pm$0.31 & \textbf{52.08}$\pm$2.40 & \textbf{78.96}$\pm$0.30 & 51.72 \\
			\bottomrule
		\end{tabular}
	\end{table}
	\begin{figure}[t]
		\centering
		\includegraphics[width=0.98\linewidth]{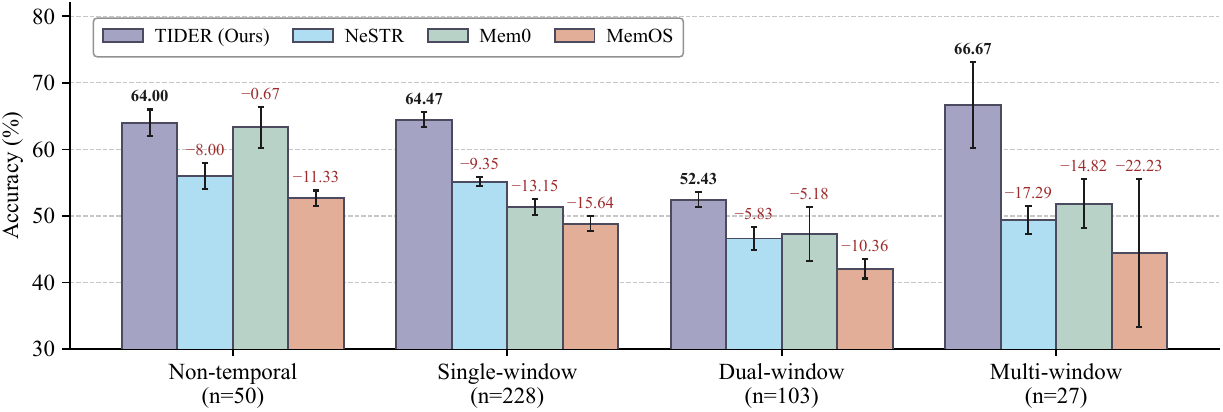}
		\caption{LifeBench accuracy stratified by gold-window count. \textbf{Bold} labels report TIDER's absolute accuracy; colored labels report each baseline's delta from TIDER within the same category.}
		\label{fig:window-stratification}
	\end{figure}
	
	\paragraph{Stratified analysis by window count.}
	To unpack overall accuracy by question type, we stratify the 408 LifeBench questions into four categories
	(Fig.~\ref{fig:window-stratification}). Time-independent queries form the \emph{non-temporal}
	category. For the remaining questions, we segment gold-evidence dates into temporal windows, where
	any gap of at least three days starts a new window, and group them into \emph{single-window},
	\emph{dual-window}, and \emph{multi-window} categories by window count. As shown in Fig.~\ref{fig:window-stratification}, TIDER
	attains the best accuracy in every category. On \emph{non-temporal} queries, its margin over the
	next-best baseline is only $0.67\%$, suggesting that closing the temporal channel does not hurt
	pure semantic recall. Across the three temporal categories, TIDER improves over the strongest
	baseline by $9.78$\% on average, with the largest gains arising when evidence must be localized across separated temporal windows (+14.8\%), suggesting that long-horizon agent memory benefits from treating temporal cues as an independent retrieval dimension rather than folding them into semantic similarity. Appendix~\ref{app:quantitative} provides more detailed analysis results with standard deviations. Appendix~\ref{app:diagnostic:cases} provides case studies from different temporal categories, illustrating how baselines can retrieve same-entity events from the wrong time, miss one of two required temporal anchors, or fall back to long-running persona facts, whereas TIDER recovers the correct evidence from long-term memory.
	
	\subsection{Ablation Study}
	\label{sec:ablation}
	
	\begin{table}[t]
		\centering
		\caption{Module ablations on LifeBench. Successively removing the distance-decay reranker, the temporal channel, and the timeline summary $\mathcal{S}_q$ from TIDER; the bottom row is a pure semantic baseline. Best in column \textbf{bold}.}
		\label{tab:ablation-modules}
		\footnotesize
		\setlength{\tabcolsep}{5.5pt}
		\begin{tabular}{lccc|cccc|cc}
			\toprule
			\textbf{Variant} & $\mathcal{S}_q$ & \textbf{Temporal} & \textbf{Decay $\gamma$} & \textbf{IE} $\uparrow$ & \textbf{MR} $\uparrow$ & \textbf{ND} $\uparrow$ & \textbf{TKU} $\uparrow$ & \textbf{Overall} $\uparrow$ & \textbf{F1} $\uparrow$ \\
			\midrule
			TIDER (full)                & \checkmark & \checkmark & \checkmark & \textbf{67.12} & \textbf{60.83} & \textbf{60.42} & \textbf{47.83} & \textbf{61.52} & \textbf{69.23} \\
			\;\;w/o decay rerank       & \checkmark & \checkmark & $\times$   & 63.70          & 58.33          & 55.21          & 41.30          & 57.60          & 67.02          \\
			\;\;w/o temporal channel   & \checkmark & $\times$   & $\times$   & 58.22          & 52.50          & 51.04          & 34.78          & 52.21          & 47.62          \\
			\;\;w/o summary (semantic) & $\times$   & $\times$   & $\times$   & 55.48          & 51.67          & 48.96          & 39.13          & 51.45          & 63.05          \\
			\bottomrule
		\end{tabular}
	\end{table}
	
	\paragraph{Module ablations.}
	Table~\ref{tab:ablation-modules} reports cumulative ablations of TIDER's main components.
	Overall accuracy decreases monotonically from $61.52\%$ to $51.45\%$ as we remove the
	distance-decay reranker, the temporal channel, and the timeline summary $\mathcal{S}_q$.
	Removing the distance-decay reranker reduces Overall accuracy by $3.92\%$, with the
	largest drop on the time-sensitive TKU category ($6.53\%$), which is especially sensitive to temporal alignment. Further removing the
	temporal channel yields an additional $5.39\%$ drop, while removing $\mathcal{S}_q$
	from the remaining semantic-only setting accounts for another $0.76\%$. 
	F1 shows a non-monotonic trend: removing the temporal channel
	reduces F1 by $19.40$ points, while further removing $\mathcal{S}_q$ recovers $15.43$
	points. This suggests that the timeline summary is most effective when paired with the
	temporal channel; without temporal constraints, it can introduce noisy context that hurts
	token-level matching. Overall, TIDER's gains arise from the coupled operation of the
	temporal interval locator, the dual-channel retriever, and the distance-decay reranker,
	rather than from any single component in isolation.
	
	To separate temporal retrieval gains from memory construction, we cross two memory representations with two retrieval methods while keeping all other settings fixed. TIDER improves Overall accuracy over semantic-only retrieval by 8.09\% on flat fixed chunks and 9.31\% on the full memory pool. Full memory construction adds 1.72\% and 2.94\% under semantic-only and TIDER retrieval, respectively. These results indicate that TIDER's gains come primarily from temporal retrieval rather than memory construction. The complete factorial comparison is provided in Appendix~\ref{app:ablation:factorial}.
	
	We also ablate the reward components while keeping the remaining training and retrieval pipeline fixed. Jointly removing the detection reward $r_\text{det}$ and window-localization reward $r_\text{win}$ reduces Overall accuracy by 5.64\% and TKU accuracy by 15.22\%. Removing the outcome reward $r_\text{out}$ or evidence reward $r_\text{evd}$ reduces Overall accuracy by 2.45\% or 1.96\%, respectively. These results show that the detection and window-localization rewards are particularly important for learning effective temporal retrieval, especially on time-sensitive questions. Detailed results are reported in Appendix~\ref{app:training:reward-loo}.
	\paragraph{Locator substitution.}
	\begin{wrapfigure}{r}{0.50\linewidth}
		\centering
		\includegraphics[width=\linewidth]{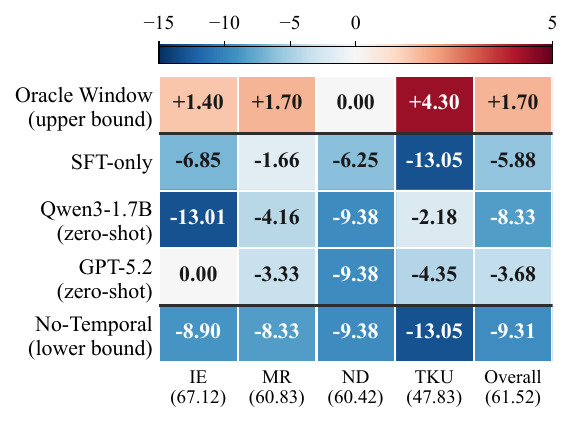}
		\caption{Locator substitution on LifeBench. Cells show accuracy differences (\%) from TIDER; column headers show TIDER's baseline scores.}
		\label{fig:locator-ablation}
	\end{wrapfigure}
	We isolate the locator by fixing the dual-channel retriever and distance-decay reranker (Fig.~\ref{fig:locator-ablation}). Removing GRPO from the Qwen3-1.7B locator drops Overall to $55.64\%$ ($-5.88\%$, with the largest loss on TKU at $-13.05\%$), while the zero-shot Qwen3-1.7B locator reaches only $53.19\%$, barely above the No-Temporal lower bound ($+0.98\%$), confirming that the temporal path requires a trained window policy. A much larger GPT-5.2 zero-shot locator improves to $57.84\%$ but remains $-3.68\%$ below TIDER, showing scale alone cannot replace reward-optimized temporal localization. Conversely, an Oracle Window upper bound ($W_\text{gold}$) adds only $+1.70\%$ Overall because TIDER's predicted windows already cover $93.77\%$ of gold dates. These results indicate the temporal path provides the structural capacity, while GRPO learns the window-alignment policy needed to activate it.

\begin{figure}[t]
		\centering
		\includegraphics[width=\linewidth]{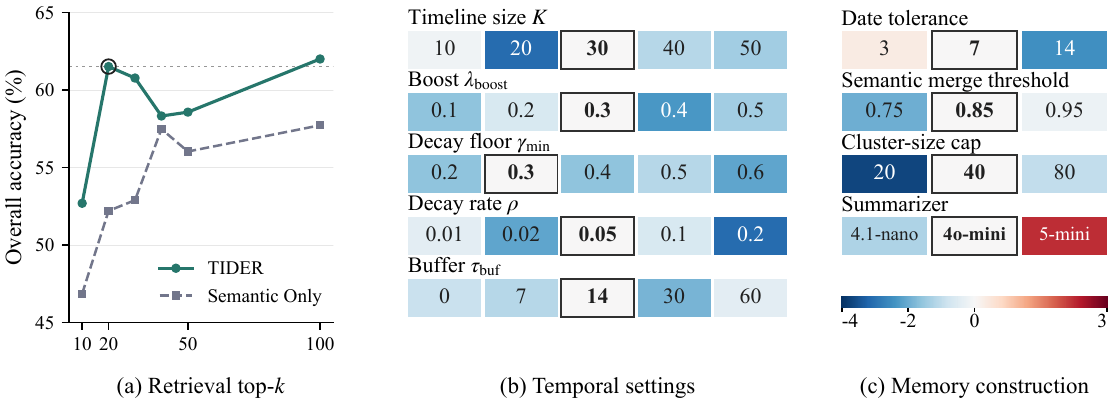}
		\caption{Hyperparameter sensitivity analysis on LifeBench.}
		\label{fig:hp-sweep}
	\end{figure}

	\paragraph{Hyperparameter robustness.}
	We examine TIDER's sensitivity to retrieval budget, temporal settings, and memory construction on LifeBench (Fig.~\ref{fig:hp-sweep}). Each sweep varies one setting while keeping the others fixed. Cells show accuracy differences(\%) from the default score, with outlines marking default settings. Detailed results are provided in Appendix~\ref{app:ablation}.

	\newpage

	\textbf{Retrieval top-$k$.} Figure~\ref{fig:hp-sweep}(a) varies the number of retrieved memories passed to the answer model. Accuracy changes non-monotonically with the retrieval budget, indicating that retrieving more memories does not consistently improve answer quality. At $k=20$, TIDER achieves $61.52\%$ accuracy, exceeding the best semantic-only result at $k=100$ by $3.79$\% while retrieving one fifth as many memories. Increasing TIDER's budget from $20$ to $100$ improves accuracy by only $0.49$\%. These results support $k=20$ as an effective balance between retrieval volume and answer accuracy.

	\textbf{Temporal settings.} Figure~\ref{fig:hp-sweep}(b) examines the timeline-input size $K$, boost coefficient, decay floor, decay rate, and temporal buffer. Here, $K$ controls the timeline information supplied to the locator, whereas $k$ in panel (a) controls the retrieved memories supplied to the answer model. Across the displayed settings, Overall accuracy ranges from $58.45\%$ to $61.52\%$. The largest within-sweep variation is $3.07\%$, with only $2.10\%$ for the decay floor and $1.86\%$ for the buffer. These results demonstrate the robustness of temporal retrieval to hyperparameter choices within the tested ranges.

	\textbf{Memory construction.} Figure~\ref{fig:hp-sweep}(c) varies the date-merge tolerance, semantic merge threshold, cluster-size cap, and summarizer model. Overall accuracy ranges from $57.84\%$ to $63.73\%$, with every tested configuration remaining above the semantic-only baseline of $52.21\%$ on the default full memory pool. The construction choices nevertheless affect performance: replacing GPT-4o-mini with GPT-5-mini raises accuracy to $63.73\%$, while reducing the cluster-size cap to $20$ items lowers it to $57.84\%$. Thus, memory construction remains an important design choice.

	\section{Conclusion}
	\label{sec:conclusion}
	
	In this work, we introduced \textbf{TIDER}, a time-aware agentic retrieval framework that addresses temporal-mismatch failures at retrieval time, where correct evidence is buried by semantically similar yet temporally mismatched candidates before any downstream reasoner can act on it. At the core of TIDER is a query-adaptive temporal interval locator, decoupled from the semantic encoder, that emits explicit temporal windows. These windows define a parallel temporal channel, whose candidates are merged with semantic candidates and reranked by temporal distance, suppressing temporally mismatched distractors while preserving semantic recall. The locator is trained via GRPO, with a composite reward that balances temporal need, window localization, evidence coverage, and answer correctness. Extensive evaluation on LifeBench and LoCoMo shows that TIDER attains state-of-the-art accuracy with only a 1.7B locator, and locator-substitution analysis confirms that the gain comes from retrieval-time temporal constraints rather than model scale. These results suggest temporal cues are crucial for pinpointing target evidence amid a vast pool of semantically similar historical events in long-term daily-life memory. We discuss limitations and broader impact in Appendix~\ref{app:impact}.

	\begin{ack}
		This work is supported by the National Key Research and Development Plan under Grant No. 2024YFB4505500 
		\& 2024YFB4505503.
	\end{ack}
	
	\small
	\bibliographystyle{plainnat}
	\bibliography{references}
	\normalsize

	
	\newpage
	\appendix
	\section*{Appendix}
	\renewcommand{\contentsname}{Appendix Contents}
	\addtocontents{toc}{\protect\setcounter{tocdepth}{2}}
	\tableofcontents
	\newpage
	\input{appendix}

	
	\newpage
	\input{checklist.tex}

\end{document}

%% file: appendix.tex


\section{Dataset Details}
\label{app:datasets}

This appendix describes the two long-horizon memory benchmarks used in our evaluation, the categories they expose, the train / evaluation splits we adopt, and the metrics under which all systems are compared.

\subsection{LifeBench}
\label{app:datasets:lifebench}
LifeBench~\citep{cheng2026lifebenchbenchmarklonghorizonmultisource} is a long-horizon multi-source personalized memory benchmark designed to stress-test memory systems beyond chat-only settings. It contains synthetic year-long personal histories for ten users (\texttt{user-0} through \texttt{user-9}), with on average roughly $13{,}000$ entries per user. Each user trace integrates multi-source daily traces---multi-session chat dialogues, phone-operation logs, trip and location histories, exercise and health-tracking records, and free-form personal notes---generated through an anonymized-persona-driven simulator with explicit temporal anchoring and cross-source consistency checks, so semantically similar events recur across the year and force the memory system to disambiguate by time and context. Questions span five categories---Information Extraction (IE), Multi-hop Reasoning (MR), Nondeclarative Memory Reasoning (ND), Temporal and Knowledge Updating (TKU), and Unanswerable (UA)---summing to $2{,}003$ items in total, and are answered open-endedly under an LLM-as-judge protocol with GPT-5-mini. State-of-the-art memory systems that score around $90\%$ on prior chat-only benchmarks (e.g., LoCoMo, LongMemEval) drop to $55\%$ overall on LifeBench, confirming that multi-source temporal personal histories pose a substantially harder retrieval and reasoning challenge. Following the LoCoMo evaluation protocol, we exclude UA, which leaves $1{,}931$ effective questions. We evaluate on the user-0 / user-1 split ($408$ questions after UA exclusion) and synthesize training samples from user-2 through user-9, so the training and evaluation users on LifeBench are disjoint.

\subsection{LoCoMo}
\label{app:datasets:locomo}
LoCoMo~\citep{maharana-etal-2024-evaluating} is a benchmark for very long-term conversational memory: each instance is a multi-session dialogue between two persona-grounded agents that, on average, spans $600$ turns and $16$K tokens across up to $32$ sessions, far exceeding the five-session horizon of prior long-term-dialogue benchmarks. The dialogues are produced by a machine--human pipeline that grounds the speakers in personas and a temporal event graph, with optional image sharing for multi-modal turns and human annotators verifying long-range consistency and event-graph faithfulness. The associated QA benchmark covers five reasoning categories---single-hop, multi-hop, temporal, open-domain, and adversarial---and was designed to expose the difficulty of comprehending long-range temporal and causal dynamics within dialogues. Following prior work, we evaluate on the $1{,}540$-question subset after excluding the adversarial category. LoCoMo serves as a zero-shot generalization test: no LoCoMo conversations or questions enter our training pipeline.
\subsection{Evaluation Metrics}
\label{app:metrics}

We evaluate all systems with the official end-to-end question-answering protocol of LoCoMo~\citep{maharana-etal-2024-evaluating} and LifeBench~\citep{cheng2026lifebenchbenchmarklonghorizonmultisource}: \textbf{Accuracy} and \textbf{F1}. This subsection gives the precise definitions and explains why we do not report classical retrieval-side metrics such as Recall@$k$ as headline numbers.

\paragraph{Accuracy.}
For each question $q$ with gold answer $a^*$ and predicted answer $\hat{a}$, we obtain a binary correctness score $\mathrm{Acc}(q)\in\{0,1\}$ via the official \texttt{locomo\_grader} judge prompt (Appendix~\ref{app:prompts:judge-out}). To suppress single-evaluation variance, we invoke the same judge prompt three independent times and adopt the majority vote ($\geq 2/3$ \textsc{correct}) as the final label, matching the $r_\text{out}$ definition used during GRPO training (Eq.~\ref{eq:r-out}). The judge model is GPT-5-mini on LifeBench and GPT-4o-mini on LoCoMo, fixed across all baselines for fairness; cross-judge variance is audited in Appendix~\ref{app:judge-consistency}. Answer generation, in contrast, is dataset-specific---LifeBench follows its official \texttt{QA\_PROMPT} and LoCoMo follows the Mem0-style \texttt{ANSWER\_PROMPT\_MEM0}, both reproduced verbatim in Appendix~\ref{app:prompts:judge-out}.

\paragraph{F1.}
F1 is the token-level harmonic mean of precision and recall between the normalised prediction $\tilde{\hat{a}}$ and the normalised gold $\tilde{a^*}$, following the standard preprocessing inherited by the LoCoMo reference implementation: lowercasing, article and punctuation removal, and white-space tokenisation. 

\paragraph{Why not Recall@$k$?}
Conventional IR metrics presuppose a canonical evidence unit shared across all systems, against which gold relevance labels can be defined uniformly. The agentic-memory baselines we compare against, as well as TIDER itself, do not consume raw conversations directly: each system ingests the raw data and emits a \emph{method-specific} memory representation through compression, extraction, or rewriting. These resulting units differ substantially in granularity---from temporally anchored event clusters in TIDER's memory pool $\mathcal{M}$ (Eq.~\ref{eq:memory-pool}), to session-level summaries in TiMem~\citep{li2026timemtemporalhierarchicalmemoryconsolidation} and MemoryBank~\citep{Zhong_Guo_Gao_Ye_Wang_2024}, per-utterance dense chunks in TsContriever~\citep{10.1145/3627673.3679800}, and temporal knowledge-graph triples in Zep~\citep{rasmussen2025zeptemporalknowledgegraph}. Consequently, top-\(k\) retrieval results are not directly comparable across systems: a top-\(k\) set of summaries covers substantially more raw evidence than a top-\(k\) set of utterances, and there is no principled normalization that aligns these heterogeneous units. Forcing a single unit, such as raw utterances, would penalize systems whose designs intentionally rewrite or condense memory, thereby conflating memory-construction choices with retrieval quality. We therefore follow the convention established by LoCoMo and LifeBench and evaluate all systems using the same downstream task performance metrics: QA accuracy and F1.

\paragraph{Retrieval Metrics over the Same Memory Pool.}
The comparability limitation above concerns systems with heterogeneous memory representations. For a controlled retrieval-side comparison, we additionally evaluate TsContriever, the semantic retriever used as NeSTR input, and TIDER on the same LifeBench memory pool $\mathcal{M}$. Because memory entries undergo extraction and rewriting, exact string matching cannot reliably align them with raw gold evidence. We therefore use an LLM coverage judge to assign binary relevance labels. Recall@$k$ is the fraction of gold evidence covered by the top-$k$ results, averaged over queries; nDCG@20 measures ranking quality using the binary relevance labels; and MRR@20 is the mean reciprocal rank of the first relevant memory within the top-20 results. These judge-based diagnostics supplement, rather than replace, the end-to-end QA metrics.

\begin{table}[htbp]
  \centering
  \footnotesize
  \setlength{\tabcolsep}{4pt}
  \caption{Retrieval metrics on the same LifeBench memory pool $\mathcal{M}$. All values are reported as percentages and use LLM-judged binary relevance. Best results are \textbf{bolded}.}
  \label{tab:shared-pool-retrieval}
  \begin{tabular}{lrrrr}
    \toprule
    Method & Recall@10 & Recall@20 & nDCG@20 & MRR@20 \\
    \midrule
    TsContriever & 17.91 & 23.84 & 20.48 & 42.98 \\
    Semantic retriever (NeSTR input) & 32.74 & 39.35 & 37.21 & 67.55 \\
    \textbf{TIDER} & \textbf{33.18} & \textbf{48.36} & \textbf{48.67} & \textbf{73.71} \\
    \bottomrule
  \end{tabular}
\end{table}

Table~\ref{tab:shared-pool-retrieval} shows that TIDER improves Recall@20 and nDCG@20 over the shared semantic retriever by $9.01$ and $11.46$ percentage points, respectively. Recall@10 is similar, while the larger Recall@20 and MRR@20 support improved evidence coverage and earlier ranking of relevant memories within the same representation. These results are conditional on the LLM coverage labels and should not be interpreted as representation-independent retrieval scores across native agentic-memory systems.

\section{Memory Pool Construction}
\label{app:memory-pool}

This appendix details the date extraction, clustering, and summarization steps that turn the raw dialogue $\mathcal{C}$ into the temporally anchored event memory pool $\mathcal{M}$ used by both retrieval channels; the full prompts driving these steps are listed in Appendix~\ref{app:prompts:cluster}.

\subsection{Date Extraction and Clustering}
\label{app:memory-pool:cluster}
Each dialogue turn is processed by an LLM that extracts atomic memory items together with their associated dates. Items are then aggregated into clusters by a temporally-bounded greedy merge: an item joins an existing cluster only when its timestamp falls within $\Delta_\text{date} = 7$ days of the cluster span and its semantic similarity (under the same dense retriever as the semantic channel) to the cluster centroid satisfies $\theta_\text{sem} \geq 0.85$; clusters are capped at $40$ items each---once a cluster reaches the cap, subsequent items spawn a new cluster even if both temporal and semantic conditions hold. Each surviving cluster $i$ records its date span $d_i = [d_i^{\text{start}}, d_i^{\text{end}}]$, an LLM-generated event summary $s_i$, and the raw items $\mathcal{I}_i$.

\subsection{Cluster Statistics}
\label{app:memory-pool:stats}
We report cluster statistics for all ten LifeBench users (EN). ``\#Items'' counts raw chat turns absorbed by all clusters; ``\#Clusters'' is the total number of LLM-summarized event clusters surviving the merge step described in Appendix~\ref{app:memory-pool}; ``Items / cluster (avg)'' is the arithmetic mean of cluster sizes; ``Timeline span (days)'' is the calendar span between the earliest and latest cluster date.

\begin{table}[t]
	\centering
	\caption{LifeBench memory-cluster statistics. ``\#Items'' counts raw turns absorbed into clusters; ``\#Clusters'' is the number of surviving event clusters after temporally aware greedy merging; ``Items / cluster (avg)'' is the arithmetic mean cluster size; ``Timeline span (days)'' is the span between the earliest and latest cluster date.}
	\label{tab:cluster-stats}
	\begin{tabular}{lrrrr}
		\toprule
		User & \#Items & \#Clusters & Items / cluster (avg) & Timeline span (days) \\
		\midrule
		\texttt{user-0} & 136{,}208 & 15{,}603 & 8.73 & 990 \\
		\texttt{user-1} & 128{,}538 & 14{,}847 & 8.66 & 833 \\
		\texttt{user-2} & 132{,}862 & 15{,}369 & 8.64 & 2{,}726 \\
		\texttt{user-3} & 131{,}961 & 15{,}243 & 8.66 & 823 \\
		\texttt{user-4} & 121{,}168 & 14{,}036 & 8.63 & 7{,}014 \\
		\texttt{user-5} & 88{,}335 & 10{,}656 & 8.29 & 1{,}308 \\
		\texttt{user-6} & 123{,}917 & 14{,}708 & 8.43 & 1{,}794 \\
		\texttt{user-7} & 20{,}683 & 3{,}136 & 6.60 & 897 \\
		\texttt{user-8} & 50{,}944 & 6{,}509 & 7.83 & 499 \\
		\texttt{user-9} & 129{,}014 & 14{,}705 & 8.77 & 1{,}071 \\
		\midrule
		Mean  & 106{,}363 & 12{,}481 & 8.52 & 1{,}795 \\
		\bottomrule
	\end{tabular}
\end{table}

\subsection{Locator-Input Timeline Coverage Audit}
\label{app:memory-pool:timeline-audit}
To check whether the clustered timeline summaries preserve the coarse temporal anchors needed by the locator, we audit the locator's actual input on the LifeBench user-0 / user-1 slice. For each question, an LLM reviewer receives the gold evidence events and the top-30 timeline summaries locator input, then judges whether those summaries contain enough coarse event-level clues to localize the evidence to the correct temporal window. 

\begin{table}[h]
	\centering
	\footnotesize
	\setlength{\tabcolsep}{8pt}
	\caption{Coarse temporal-window localization coverage of the locator's top-30 timeline-summary input on LifeBench. Coverage is judged by an LLM reviewer and averaged per question.}
	\label{tab:locator-top30-localization}
	\begin{tabular}{lcc}
		\toprule
		\textbf{Category} & \textbf{N} & \textbf{Top-30 localization coverage} \\
		\midrule
		Non-temporal  & 50  & 87.60 \\
		Single-window & 228 & 89.45 \\
		Dual-window   & 103 & 73.51 \\
		Multi-window  & 27  & 78.85 \\
		\midrule
		Overall       & 408 & 84.51 \\
		\bottomrule
	\end{tabular}
\end{table}
\section{Locator Training Details}
\label{app:training}

This appendix specifies the training procedure for the locator $\pi_\theta$: the SFT cold-start, the GRPO online phase, the composite-reward weighting, and the difficulty-stratified rollout.

\subsection{Training Data Construction}
\label{app:training:data}
We synthesize training samples from LifeBench user-2 through user-9. For each query $q$, we annotate four gold supervisions:
\begin{itemize}\itemsep1pt
	\item $\texttt{gold\_temporal} \in \{\text{true}, \text{false}\}$, indicating whether the query is time-sensitive.
	\item the gold window set $W_\text{gold}$, present only when $\texttt{gold\_temporal} = \text{true}$.
	\item the gold evidence $\mathcal{E}^* = \{e_k\}_{k=1}^{|\mathcal{E}^*|}$.
	\item the gold answer $a^*$.
\end{itemize}
\paragraph{Annotation provenance.} The four supervisions above originate from heterogeneous sources rather than a single annotation pass.

\textit{Gold answer $a^\ast$.} Each $a^\ast$ is taken verbatim from LifeBench, which ships an authoritative answer for every QA instance: a single option letter (A/B/C/D) for multiple-choice questions and a short reference text for open-ended questions. No filtering or rewriting is applied.

\textit{Gold evidence $\mathcal{E}^\ast$.} Each LifeBench QA instance carries an \texttt{evidence\_resolved} field that lists one record per supporting evidence, with source identifier, type, date, and raw texts. We adopt the field verbatim and convert each record into a $(\text{date}, \text{text})$ tuple consumed by the evidence reward $r_\text{evd}$.

\textit{Gold windows $W_\text{gold}$.} Derived deterministically from the dates in $\mathcal{E}^\ast$. Each evidence date is first cast as a degenerate single-day interval $[d, d]$; intervals are then sorted by start date and merged whenever the gap to the previous interval is fewer than three calendar days, otherwise emitted as a new window. Evidence with missing or unparseable dates is skipped. The resulting $W_\text{gold}$ collapses dense evidence clusters into a single tight window and exposes sparse clusters as multi-window targets, providing the natural difficulty stratification used in Appendix~\ref{app:training}.

\textit{Temporal flag $\texttt{gold\_temporal}$.} A binary indicator of whether $q$ is time-sensitive. We obtain an initial label by prompting GPT-5.2 to classify each query in isolation, and a human annotator subsequently verifies every label and overrides the model decision when warranted. This two-stage pipeline reduces both false positives (queries whose evidence happens to carry incidental dates) and false negatives (implicitly temporal queries whose surface form omits explicit time markers).

The resulting set contains $1{,}589$ samples spanning the four window-count buckets used by the difficulty-stratified rollout in Appendix~\ref{app:training}: $38$ non-temporal ($|W_\text{gold}|=0$, $2.4\%$), $1{,}014$ single-window ($|W_\text{gold}|=1$, $63.8\%$), $383$ dual-window ($|W_\text{gold}|=2$, $24.1\%$), and $154$ multi-window ($|W_\text{gold}|\geq 3$, $9.7\%$).

\subsection{SFT Cold-Start}
\label{app:training:sft}
SFT aligns $\pi_\theta$ to the locator output schema (Appendix~\ref{app:prompts}) and seeds an initial window guess. Training data are pairs $(q, z^*)$ derived from the gold annotations. We initialize from Qwen3-1.7B--and train for $2$ epochs with peak learning rate $2\times 10^{-5}$, per-GPU batch size $4$, gradient-accumulation steps $2$, and maximum sequence length $6144$ tokens, on $2$ GPUs; the effective batch size is therefore $16$.

\subsection{GRPO Hyperparameters}
\label{app:training:grpo}
\begin{table}[t]
	\centering
	\caption{GRPO hyperparameters for training the locator $\pi_\theta$. The SFT-aligned Qwen3-1.7B policy is further optimized with GRPO using the configuration below.}
	\label{tab:grpo-hparams}
	\begin{tabular}{lr}
		\toprule
		Parameter & Value \\
		\midrule
		Advantage estimator & GRPO \\
		KL coefficient $\beta$ (loss) & 0.1 \\
		Group size $G$ (rollouts per prompt) & 8 \\
		Rollout temperature & 0.7 \\
		Peak learning rate & $5\times10^{-6}$ \\
		LR schedule & cosine, warmup ratio 0.1 \\
		PPO epochs per batch & 1 \\
		Train / mini batch size & 16 / 16 \\
		Micro batch size per GPU & 8 \\
		Total epochs & 6 \\
		$R_\text{min}$ for unparseable rollouts & $0$ \\
		\bottomrule
	\end{tabular}
\end{table}
For the GRPO objective in Eq.~\eqref{eq:grpo}, the likelihood ratio for rollout $j$ is
\[
\rho^{(j)} =
\frac{\pi_\theta(z^{(j)} \mid q, \mathcal{S}_q)}
{\pi_{\theta_\text{old}}(z^{(j)} \mid q, \mathcal{S}_q)},
\]
and $A^{(j)}$ is the group-normalized advantage computed from the composite rewards of the $G$ rollouts sampled for the same query. This normalization follows GRPO: the rewards within a group are centered and scaled before clipping, while the KL term is applied against the reference policy $\pi_\text{ref}$.

\subsection{Reward Weights}
\label{app:training:reward-weights}
The composite reward (Eq.~\eqref{eq:reward}) uses weights
\[
w_\text{det} = 0.10, \quad
w_\text{win} = 0.25, \quad
w_\text{evd} = 0.20, \quad
w_\text{out} = 0.45,
\qquad \textstyle\sum_\bullet w_\bullet = 1.
\]
The end-task signal $w_\text{out}$ takes the largest share because answer correctness is the only term that depends jointly on detection, window prediction, and downstream synthesis; the three intermediate weights $w_\text{det}, w_\text{win}, w_\text{evd}$ supply dense supervision early in training, when $r_\text{out}$ is still sparse. Among the intermediate signals, $w_\text{win} > w_\text{evd}$ because window-prediction errors propagate to evidence retrieval, so localizing the window is the higher-leverage upstream signal.

For $r_\text{win}$ in Eq.~\eqref{eq:r-win}, the bidirectional IoU score is computed only for temporal samples with nonempty $W_\text{gold}$ and nonempty $W_\text{pred}$. Let $\mathcal{T}_\text{win}$ denote the case where $\texttt{gold\_temporal}=\text{true}$, $W_\text{gold}\neq\emptyset$, and $W_\text{pred}\neq\emptyset$. The full boundary-condition form is:
\begin{align*}
	b(W_\text{pred}, W_\text{gold})
	&= \tfrac{1}{2}\!\left(
	\tfrac{1}{|W_\text{gold}|}\!\sum_{G \in W_\text{gold}} \max_{P \in W_\text{pred}} \varphi(P, G)
	+ \tfrac{1}{|W_\text{pred}|}\!\sum_{P \in W_\text{pred}} \max_{G \in W_\text{gold}} \varphi(P, G)
	\right), \\
	r_\text{win}
	&= \begin{cases}
		b(W_\text{pred}, W_\text{gold}), & \mathcal{T}_\text{win}, \\
		0, & \text{otherwise}.
	\end{cases}
\end{align*}
Non-temporal samples, empty predictions, and samples whose gold evidence yields no parseable temporal window therefore receive $r_\text{win}=0$. This convention avoids undefined maxima or normalizers and leaves the decision of whether the temporal channel should be opened to $r_\text{det}$.

\subsection{Difficulty-Stratified Rollout and \texorpdfstring{$R_\text{min}$}{R\_min}}
\label{app:training:rollout}
Rollouts are reweighted by an asymmetric U-shape difficulty-aware reweighting (DAR) schedule indexed by $|W_\text{gold}|$: the two extremes of the difficulty axis are up-weighted, while the easier middle buckets are kept at the base weight $1.0$. The factor for non-temporal questions ($\times 10$) is much larger than that for multi-window questions ($\times 4$) because the non-temporal bucket is severely under-represented in the training distribution ($2.4\%$ of samples; see Appendix~\ref{app:training}), whereas the multi-window bucket, while still skewed, is less so ($9.7\%$). Concretely:
\begin{table}[t]
	\centering
	\caption{Difficulty-aware rollout reweighting (DAR) by gold-window count. ``Effective probability'' applies the DAR factors to the empirical training distribution and renormalizes the result.}
	\label{tab:dar}
	\begin{tabular}{lrr}
		\toprule
		Bucket ($|W_\text{gold}|$) & DAR factor & Effective probability \\
		\midrule
		0 (non-temporal) & 10.0 & 0.159 \\
		1 (single-window) & 1.0 & 0.424 \\
		2 (dual-window) & 1.0 & 0.160 \\
		$\geq 3$ (multi-window) & 4.0 & 0.257 \\
		\bottomrule
	\end{tabular}
\end{table}
The effective-probability column applies the DAR factors to the empirical training distribution reported in Appendix~\ref{app:training} (per-bucket sample counts $38 / 1{,}014 / 383 / 154$), then renormalizes. Rollouts whose locator output cannot be parsed by the JSON schema (Appendix~\ref{app:prompts}) are clamped to $R = R_\text{min} = 0$ and bypass the LLM judge.

\subsection{Reward-Component Leave-One-Out Ablation}
\label{app:training:reward-loo}
Table~\ref{tab:reward-loo} reports leave-one-out ablations over the four composite-reward terms of Eq.~\eqref{eq:reward}. Each non-full row removes the indicated reward term during the GRPO phase while keeping the SFT cold-start, the locator backbone, the dual-channel retriever, and the downstream LLM fixed. Detection and window-IoU are removed jointly because $r_\text{win}$ is gated on $r_\text{det}{=}1$.

\begin{table}[h]
	\centering
	\caption{Reward-component leave-one-out ablations on LifeBench. The first row is full TIDER; subsequent rows remove the indicated term(s) during GRPO. Best results are \textbf{bolded}.}
	\label{tab:reward-loo}
	\footnotesize
	\setlength{\tabcolsep}{4pt}
	\begin{tabular}{l|cccc|cc}
		\toprule
		\textbf{Variant} & \textbf{IE} $\uparrow$ & \textbf{MR} $\uparrow$ & \textbf{ND} $\uparrow$ & \textbf{TKU} $\uparrow$ & \textbf{Overall} $\uparrow$ & \textbf{F1} $\uparrow$ \\
		\midrule
		\textbf{TIDER (SFT+GRPO, 1.7B)} & \textbf{67.12} & \textbf{60.83} & \textbf{60.42} & \textbf{47.83} & \textbf{61.52} & \textbf{69.23} \\
		\;\;w/o $r_\text{win}, r_\text{det}$ & 60.96 & 56.67 & 58.33 & 32.61 & 55.88 & 67.46 \\
		\;\;w/o $r_\text{out}$               & 66.44 & 56.67 & 58.33 & 43.48 & 59.07 & 69.20 \\
		\;\;w/o $r_\text{evd}$                  & 64.38   & 60.83   & 56.25   & 47.83   & 59.56 & 68.81   \\
		\bottomrule
	\end{tabular}
\end{table}

\subsection{Training Curves}
\label{app:training:curves}
\begin{figure}[h]
	\centering
	\includegraphics[width=0.85\linewidth]{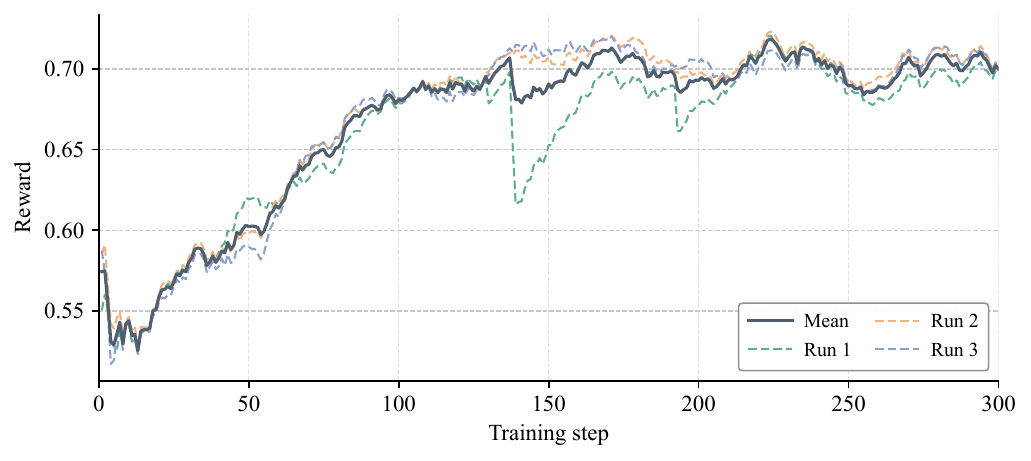}
	\caption{GRPO training curves of the 1.7B locator on LifeBench.}
	\label{fig:training-curves}
\end{figure}

Figure~\ref{fig:training-curves} plots the per-step reward for three independent GRPO runs that share an identical training recipe and differ only in launch timing. Each dashed curve is one run's first 300 steps after EMA smoothing ($\alpha{=}0.95$, bias-corrected); the solid line is the per-step mean across the  runs---averaging all three for step. The three trajectories are visually superimposed during the early ascent and remain within a narrow band at the plateau, indicating that the GRPO gain reported in \S\ref{sec:results} is reproducible rather than an artifact of a single training trajectory.

\subsection{Lightweight Locator: 1.7B vs 4B}
\label{app:training:capacity}
\begin{figure}[h]
	\centering
	\includegraphics[width=0.85\linewidth]{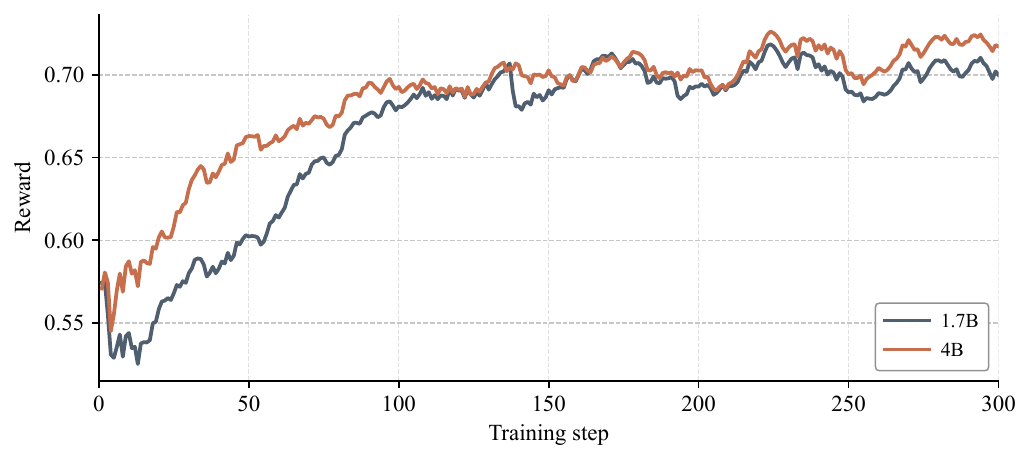}
	\caption{Capacity ablation on the GRPO training curve. Per-step reward (EMA $\alpha{=}0.95$, bias-corrected) for the 4B locator and the 1.7B.}
	\label{fig:capacity-curves}
\end{figure}

\textit{Locator capacity is not the source of TIDER's gains.} We re-ran the full TIDER pipeline with a 4B locator under an identical training recipe; only the SFT cold-start and GRPO traces are model-specific, while the dual-channel retriever, prompts, and downstream LLM are unchanged. Figure~\ref{fig:capacity-curves} overlays the smoothed reward of the 4B run on the 1.7B run from Figure~\ref{fig:training-curves}: the two trajectories rise on a similar shape and converge to plateaus differing by less than $0.02$ in reward, with the 4B run ascending marginally faster in the early phase. On the evaluation side, Table~\ref{tab:capacity-ablation} reports end-to-end accuracy on LifeBench. Halving the locator from $4$B to $1.7$B parameters leaves Overall accuracy and average F1 effectively unchanged ($61.52$ vs $61.03$; $0.6923$ vs $0.6957$), with per-category swings within $\pm 2.2$ pp. We therefore adopt the $1.7$B locator as the default, attributing TIDER's headroom to the dual-channel temporal anchor rather than to locator scale. TIDER thus inherits a lightweight characteristic, achieving competitive end-to-end accuracy under a substantially smaller locator parameter budget.

\begin{table}[h]
	\centering
	\caption{Locator capacity ablation on LifeBench (user-0/1, $n{=}408$). Per-category accuracy is reported alongside Overall accuracy and average F1. Per-column best results are \textbf{bolded}; the $1.7$B locator is our adopted default.}
	\label{tab:capacity-ablation}
	\footnotesize
	\setlength{\tabcolsep}{4pt}
	\begin{tabular}{l|cccc|cc}
		\toprule
		\textbf{Locator} & \textbf{IE} $\uparrow$ & \textbf{MR} $\uparrow$ & \textbf{ND} $\uparrow$ & \textbf{TKU} $\uparrow$ & \textbf{Overall} $\uparrow$ & \textbf{F1} $\uparrow$ \\
		\midrule
		TIDER (4B)              & 65.07          & \textbf{61.67} & 59.38          & \textbf{50.00} & 61.03          & \textbf{69.57} \\
		\textbf{TIDER (1.7B)}   & \textbf{67.12} & 60.83          & \textbf{60.42} & 47.83          & \textbf{61.52} & 69.23          \\
		$\Delta$ ($1.7$B$-4$B)  & $+2.05$        & $-0.84$        & $+1.04$        & $-2.17$        & $+0.49$        & $-0.34$        \\
		\bottomrule
	\end{tabular}
\end{table}

\section{Ablation Study}
\label{app:ablation}

\subsection{Hyperparameter Robustness}
\label{app:hp-sweep}

\begin{figure}[t]
	\centering
	\includegraphics[width=0.95\linewidth]{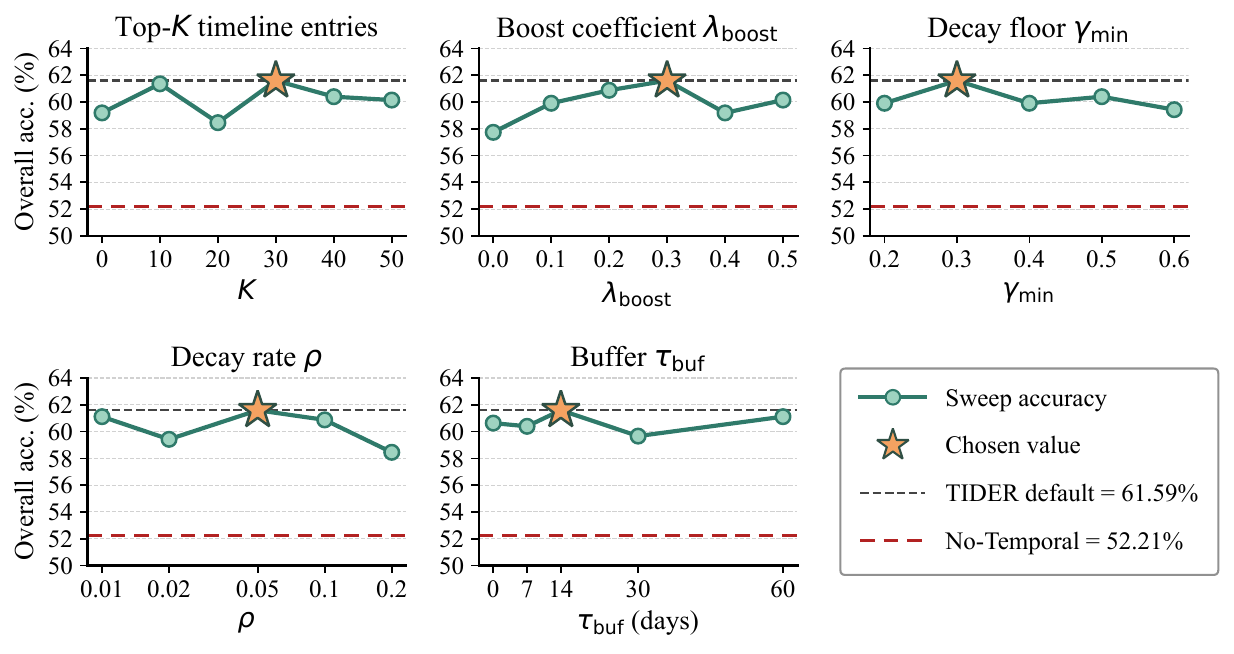}
	\caption{Hyperparameter robustness on LifeBench. Five hyperparameters are swept individually around the default configuration ($K{=}30$, $\lambda_\text{boost}{=}0.3$, $\gamma_\text{min}{=}0.3$, $\rho{=}0.05$, $\tau_\text{buf}{=}14$). Each panel shows Overall accuracy (\%) versus the swept value; the orange star marks the chosen default; the upper dashed line is TIDER's default-config score; the lower dotted line is the No-Temporal lower bound.}
	\label{fig:hp-sweep-app}
\end{figure}

This subsection reports the full hyperparameter sweep referenced in \S\ref{sec:ablation}.
Across the $25$ configurations summarized in Fig.~\ref{fig:hp-sweep-app}, Overall accuracy stays within a narrow $[57.73, 61.52]\%$ band, far above the No-Temporal bound ($52.21\%$). Per-hyperparameter ranges are: $\lambda_\text{boost}$ ($3.86\%$, most sensitive), top-$K$ timeline entries ($3.14\%$), decay rate $\rho$ ($3.14\%$), floor $\gamma_\text{min}$ ($2.10\%$), buffer $\tau_\text{buf}$ ($1.86\%$, most stable). The chosen defaults sit at or near each per-panel optimum, and a single shared configuration applies across both LifeBench evaluation users without per-user tuning.

\subsection{Memory Construction--Retrieval Factorial Comparison}
\label{app:ablation:factorial}
To separate the effects of memory construction and retrieval, we cross two memory representations with two retrieval methods on LifeBench. The fixed-chunk representation uses 512-token chunks with timestamps inherited from their source sessions, without LLM-based extraction, clustering, rewriting, or summarization; the locator directly consumes the timestamped chunks. The full representation uses the temporally anchored memory pool $\mathcal{M}$. Within each representation, we compare semantic-only retrieval with TIDER retrieval while holding the downstream answer model and retrieval budget fixed.

\begin{table}[htbp]
  \centering
  \footnotesize
  \setlength{\tabcolsep}{6pt}
  \caption{A $2\times2$ factorial comparison of memory representation and retrieval method on LifeBench. Values are Overall accuracy (\%).}
  \label{tab:construction-retrieval-factorial}
  \begin{tabular}{lrr}
    \toprule
    Memory representation & Semantic retrieval & TIDER retrieval \\
    \midrule
    Fixed chunks + session timestamps & 50.49 & 58.58 \\
    Full temporally anchored memory pool $\mathcal{M}$ & 52.21 & 61.52 \\
    \bottomrule
  \end{tabular}
\end{table}

TIDER improves accuracy by $8.09$ percentage points on fixed chunks and $9.31$ points on the full memory pool. Memory construction adds $1.72$ points under semantic retrieval and $2.94$ points under TIDER retrieval. The retrieval-time mechanism therefore remains effective without LLM-based memory construction, while the full memory pool provides an additional, retrieval-dependent benefit.

\subsection{Retrieval Top-$k$ Sweep}
\label{app:ablation:retrieval-k}
We vary the number of retrieved memory units passed to the answer generator, $k$, while fixing the memory pool, answer generator, and all other settings. This retrieval budget differs from the top-$K$ timeline-input size in Appendix~\ref{app:hp-sweep}, which controls the information supplied to the locator.

\begin{table}[htbp]
  \centering
  \footnotesize
  \setlength{\tabcolsep}{7pt}
  \caption{Retrieval-budget sweep on LifeBench with the same memory pool and answer generator. Values are Overall accuracy (\%).}
  \label{tab:retrieval-k-sweep}
  \begin{tabular}{lrrrrrr}
    \toprule
    Retrieval budget $k$ & 10 & 20 & 30 & 40 & 50 & 100 \\
    \midrule
    TIDER & 52.70 & 61.52 & 60.78 & 58.33 & 58.58 & 62.01 \\
    Semantic Only & 46.86 & 52.21 & 52.90 & 57.49 & 56.04 & 57.73 \\
    \bottomrule
  \end{tabular}
\end{table}

Accuracy varies non-monotonically with $k$. Increasing the semantic retrieval budget narrows the gap, but the best semantic-only result within the tested range is $57.73\%$ at $k=100$, still $3.79$ percentage points below TIDER at $k=20$ ($61.52\%$). Increasing TIDER's budget from $20$ to $100$ improves accuracy by only $0.49$ points. These results support $k=20$ as an effective operating point: TIDER achieves higher accuracy than the best tested semantic-only setting while retrieving one fifth as many memory units. This comparison concerns memory-unit counts, not measured token costs or end-to-end latency.

\subsection{Memory-Construction Hyperparameter Sweep}
\label{app:ablation:construction-sweep}
We vary the date-merge tolerance $\Delta_\text{date}$, semantic merge threshold $\theta_\text{sem}$, cluster-size cap, and summarizer model individually, keeping the other settings at their defaults. The default construction uses $\Delta_\text{date}=7$ days, $\theta_\text{sem}=0.85$, a cap of $40$ items, and GPT-4o-mini as the summarizer. Table~\ref{tab:construction-sweep} reports the corresponding LifeBench Overall accuracies.

\begin{table}[htbp]
  \centering
  \footnotesize
  \setlength{\tabcolsep}{4pt}
  \caption{Memory-construction sensitivity on LifeBench. Settings and accuracies in each row correspond in order; accuracy is reported in percent.}
  \label{tab:construction-sweep}
  \begin{tabular}{lp{0.46\linewidth}l}
    \toprule
    Parameter & Settings & Accuracy (\%) \\
    \midrule
    $\Delta_\text{date}$ (days) & 3 / 7 / 14 & 61.76 / 61.52 / 59.07 \\
    $\theta_\text{sem}$ & 0.75 / 0.85 / 0.95 & 59.56 / 61.52 / 61.27 \\
    Cluster-size cap & 20 / 40 / 80 & 57.84 / 61.52 / 60.54 \\
    Summarizer & GPT-4.1-nano / GPT-4o-mini / GPT-5-mini & 60.29 / 61.52 / 63.73 \\
    \bottomrule
  \end{tabular}
\end{table}

Accuracy ranges from $59.07\%$ to $61.76\%$ for $\Delta_\text{date}$, $59.56\%$ to $61.52\%$ for $\theta_\text{sem}$, $57.84\%$ to $61.52\%$ for the cap, and $60.29\%$ to $63.73\%$ across summarizer models. All tested configurations remain above the default full-pool semantic-only accuracy of $52.21\%$. This comparison supports robustness relative to that reference operating point.

\section{Baseline Details}
\label{app:baselines}

This appendix expands the one-line baseline descriptions in \S\ref{sec:setup} and details the implementation choices that ensure a fair comparison.

\subsection{Method Summaries}
\label{app:baselines:methods}
We group the six baselines into three families along their orthogonal design dimensions: agentic memory frameworks (MemOS, TiMem, Zep, Mem0) that target memory organization and lifecycle management, time-aware retrieval (TsContriever) that targets the retrieval encoder, and temporal reasoning (NeSTR) that targets answer-time inference over retrieved evidence. The summaries below paraphrase each paper's stated contribution; the closing line marks what we preserve from the original method versus what we adapt to keep the comparison on equal footing with TIDER.

\textbf{MemOS}~\citep{li2025memosoperatingmemoryaugmentedgeneration} treats memory as a manageable system resource and proposes a memory operating system that unifies the representation, scheduling, and evolution of plaintext, activation-based, and parameter-level memories. Its basic unit, the MemCube, encapsulates memory content together with provenance and versioning metadata; MemCubes can be composed, migrated, and fused over time, bridging retrieval-style memory with parameter-based learning under explicit lifecycle control. \emph{Adapted:} the downstream answer model is replaced by our shared LLM (GPT-5-mini on LifeBench, GPT-4o-mini on LoCoMo) so that gains attributable to memory organization are isolated from the choice of generator.

\textbf{TiMem}~\citep{li2026timemtemporalhierarchicalmemoryconsolidation} is a temporal--hierarchical memory framework that organizes conversations through a Temporal Memory Tree (TMT), enabling systematic memory consolidation from raw observations to progressively abstracted persona representations. It is characterized by three properties: temporal--hierarchical organization through TMT, semantic-guided consolidation that integrates memory across hierarchical levels without fine-tuning, and complexity-aware memory recall that balances precision and efficiency across queries of varying complexity. \emph{Adapted:} the downstream answer LLM is unified with the rest of the table, leaving TiMem's memory tree as the only varying component.

\textbf{Zep}~\citep{rasmussen2025zeptemporalknowledgegraph} is a memory-layer service for AI agents whose core component, Graphiti, dynamically synthesizes unstructured conversational data and structured business data into a temporally-aware knowledge graph that maintains historical relationships with explicit validity intervals. Zep was originally evaluated on Deep Memory Retrieval (DMR) and LongMemEval, where it improves over MemGPT and shows substantial gains on cross-session synthesis and long-term context maintenance with reduced response latency. \emph{Adapted:} the answer model is unified across baselines.

\textbf{Mem0}~\citep{chhikara2025mem0buildingproductionreadyai} is a scalable memory-centric architecture that dynamically extracts, consolidates, and retrieves salient information from ongoing multi-session conversations, with an enhanced graph-based variant that captures relational structure among conversational elements. On the LOCOMO benchmark Mem0 reports a $26\%$ relative improvement over an OpenAI memory baseline under LLM-as-Judge while reducing p95 latency by $91\%$ and saving more than $90\%$ of token cost. \emph{Adapted:} the downstream answer LLM and top-$k$ retrieval budget are unified.

\textbf{TsContriever}~\citep{10.1145/3627673.3679800} (a.k.a.\ TS-Retriever) is a time-sensitive retrieval-augmented generation method built on top of Contriever. It augments dense retrieval with supervised contrastive learning using temporally-constrained negatives, performs query-side fine-tuning to make the encoder sensitive to explicit temporal expressions, and routes queries through a temporal-need classifier so that general queries fall back to standard semantic retrieval. The original paper reports clear gains on a custom time-sensitive QA benchmark while preserving general-query effectiveness.  \emph{Adapted:} We preserve the four official checkpoints (plain, document, query, router) and the routing classifier. We disable the TIDER locator and plug TsContriever into our shared memory pool, retrieval pipeline, top-$k$, and answer LLM, so the difference reflects the encoder change alone.

\textbf{NeSTR}~\citep{Liang_Zeng_Zhao_Zhao_2026} is a Neuro-Symbolic Temporal Reasoning framework that integrates structured symbolic representations of temporal relations with hybrid reflective reasoning to enhance LLMs' temporal sensitivity. It explicitly preserves temporal relations through symbolic encoding, enforces logical consistency through verification, and corrects flawed inferences via abductive reflection, achieving zero-shot gains on temporal QA benchmarks without any fine-tuning.  \emph{Adapted:} We preserve the NeSTR  for response generation, including symbolic temporal encoding, consistency verification, and abductive reflection. We disable TIDER's locator, temporal grounding and temporal reranker at the search stage, share the memory pool and dense retriever.

\subsection{Excluded Baselines}
\label{app:baselines:excluded}
We do not include training-based temporal-reasoning methods such as Memory-T1~\citep{du2026memoryt} as direct baselines, since they fine-tune the response model whereas our evaluation keeps the downstream LLM fixed across methods.

\section{Detailed Quantitative Results}
\label{app:quantitative}

This appendix reports the detailed quantitative evidence behind the LifeBench comparisons: accuracy stratified by gold-window-count category (\S\ref{app:quantitative:category-acc}) and paired significance tests on per-question correctness (\S\ref{app:quantitative:significance}).

\subsection{Stratified Accuracy by Category}
\label{app:quantitative:category-acc}

Table~\ref{tab:category-acc} reports per-category accuracy of all seven baselines on the official LifeBench user-0 / user-1 slice ($n{=}408$), using the same four categories visualized in Fig.~\ref{fig:window-stratification}: non-temporal ($n{=}50$), single-window ($n{=}228$), dual-window ($n{=}103$), and multi-window ($n{=}27$). Means and standard deviations are reported over three independent evaluation runs.

\begin{table}[h]
	\centering
	\footnotesize
	\setlength{\tabcolsep}{4pt}
	\caption{LifeBench accuracy stratified by gold-window-count category. Best results are \textbf{bolded}; \underline{underline} indicates second best.}
	\label{tab:category-acc}
	\begin{tabular}{l|cccc|c}
		\toprule
		\textbf{Method} & \textbf{Non-temporal} & \textbf{Single-window} & \textbf{Dual-window} & \textbf{Multi-window} & \textbf{Overall} \\
		\midrule
		MemOS        & 52.67$\pm$1.15             & 48.83$\pm$1.10             & 42.07$\pm$1.48             & 44.44$\pm$11.11            & 47.30$\pm$0.88 \\
		TiMem        & 23.33$\pm$1.15             & 16.96$\pm$1.10             & 17.80$\pm$1.48             & 13.58$\pm$4.28             & 17.73$\pm$0.57 \\
		Zep          & 48.67$\pm$7.02             & 38.74$\pm$0.67             & 33.98$\pm$1.68             & 33.33$\pm$0.00             & 38.40$\pm$0.37 \\
		Mem0         & \underline{63.33}$\pm$3.06 & 51.32$\pm$1.16             & \underline{47.25}$\pm$4.04 & \underline{51.85}$\pm$3.70 & 51.80$\pm$1.26 \\
		TsContriever & 38.00$\pm$2.00             & 28.07$\pm$1.16             & 23.62$\pm$1.12             & 13.58$\pm$4.28             & 27.21$\pm$0.42 \\
		NeSTR        & 56.00$\pm$2.00             & \underline{55.12}$\pm$0.67 & 46.60$\pm$1.68             & 49.38$\pm$2.14             & \underline{52.70}$\pm$0.65 \\
		\midrule
		\rowcolor{gray!20}
		\textbf{TIDER} & \textbf{64.00}$\pm$2.00 & \textbf{64.47}$\pm$1.16 & \textbf{52.43}$\pm$1.12 & \textbf{66.67}$\pm$6.42 & \textbf{61.52}$\pm$0.86 \\
		\bottomrule
	\end{tabular}
\end{table}

\subsection{Paired Significance Tests}
\label{app:quantitative:significance}

Table~\ref{tab:paired-significance} reports paired significance tests comparing TIDER against each baseline on the same 408 answerable LifeBench questions. We use the two-sided exact McNemar test on paired binary correctness outcomes and apply Holm correction across the six TIDER-versus-baseline comparisons. In the table, $b$ counts questions answered correctly by TIDER but incorrectly by the baseline, while $c$ counts the reverse direction.

\begin{table}[h]
	\centering
	\footnotesize
	\setlength{\tabcolsep}{6pt}
	\caption{Paired significance tests for TIDER against each baseline on LifeBench. $\Delta$Acc is reported in percentage points.}
	\label{tab:paired-significance}
	\begin{tabular}{lrrrrr}
		\toprule
		\textbf{Comparison} & \textbf{$\Delta$Acc} & \textbf{$b$} & \textbf{$c$} & \textbf{Exact $p$} & \textbf{Holm $p$} \\
		\midrule
		TIDER vs.\ MemOS        & 13.48 &  92 & 37 & $1.40{\times}10^{-6}$  & $4.19{\times}10^{-6}$ \\
		TIDER vs.\ TiMem        & 43.14 & 188 & 12 & $8.11{\times}10^{-42}$ & $4.87{\times}10^{-41}$ \\
		TIDER vs.\ Zep          & 22.79 & 121 & 28 & $5.59{\times}10^{-15}$ & $2.24{\times}10^{-14}$ \\
		TIDER vs.\ Mem0         &  8.33 &  66 & 32 & $7.70{\times}10^{-4}$  & $8.94{\times}10^{-4}$ \\
		TIDER vs.\ TsContriever & 34.07 & 149 & 10 & $6.25{\times}10^{-33}$ & $3.12{\times}10^{-32}$ \\
		TIDER vs.\ NeSTR        &  8.09 &  59 & 26 & $4.47{\times}10^{-4}$  & $8.94{\times}10^{-4}$ \\
		\bottomrule
	\end{tabular}
\end{table}

All TIDER-versus-baseline differences remain significant after Holm correction, including the closest comparisons against Mem0 and NeSTR.

\section{Diagnostic Analysis and Case Studies}
\label{app:diagnostic}

This appendix analyzes why methods fail on LifeBench. We first summarize LLM-judged error causes over incorrect answers (\S\ref{app:diagnostic:error-cause}), then provide qualitative case studies covering single-, dual-, and multi-window queries (\S\ref{app:diagnostic:cases}).
\subsection{Error-Cause Analysis}
\label{app:diagnostic:error-cause}

For each incorrect answer, we ask an LLM reviewer to assign exactly one primary error cause by comparing the question, gold answer, model answer, gold evidence, and retrieved context (Appendix~\ref{app:prompts:error-cause}). The labels are: \emph{retrieval miss}, where the answer-bearing evidence is absent; \emph{temporal mismatch}, where related evidence is retrieved from the wrong time; \emph{partial evidence}, where only part of the required evidence is retrieved; and \emph{reason failure}, where sufficient evidence is present but the answer model reasons or selects incorrectly. We define \emph{retrieval-side failure} as the union of retrieval miss, temporal mismatch, and partial evidence.

\begin{table}[h]
	\centering
	\scriptsize
	\setlength{\tabcolsep}{3.5pt}
	\caption{LLM-judged primary error causes over incorrect LifeBench answers. Percentages are computed within each method's incorrect answers.}
	\label{tab:error-cause-analysis}
	\resizebox{\linewidth}{!}{
		\begin{tabular}{lrrrrrr}
			\toprule
			\textbf{Method} & \textbf{Wrong} & \textbf{Retrieval miss} & \textbf{Temporal mismatch} & \textbf{Partial evidence} & \textbf{Reason failure} & \textbf{Retrieval-side failure} \\
			\midrule
			TIDER        & 157 &  27 (17.2\%) & 107 (68.2\%) &  5 (3.2\%) & 18 (11.5\%) & 139 (88.5\%) \\
			MemOS        & 212 &  41 (19.3\%) & 132 (62.3\%) & 13 (6.1\%) & 26 (12.3\%) & 186 (87.7\%) \\
			TiMem        & 333 & 174 (52.3\%) & 142 (42.6\%) &  7 (2.1\%) & 10 (3.0\%)  & 323 (97.0\%) \\
			Zep          & 250 &  88 (35.2\%) & 115 (46.0\%) & 18 (7.2\%) & 29 (11.6\%) & 221 (88.4\%) \\
			Mem0         & 191 &  45 (23.6\%) & 123 (64.4\%) &  9 (4.7\%) & 14 (7.3\%)  & 177 (92.7\%) \\
			TsContriever & 296 & 143 (48.3\%) & 122 (41.2\%) &  8 (2.7\%) & 23 (7.8\%)  & 273 (92.2\%) \\
			NeSTR        & 190 &  38 (20.0\%) & 119 (62.6\%) &  2 (1.1\%) & 31 (16.3\%) & 159 (83.7\%) \\
			\bottomrule
	\end{tabular}}
\end{table}

Table~\ref{tab:error-cause-analysis} shows that most failures are retrieval-side rather than pure answer-generation failures. Temporal mismatch is the dominant failure type for MemOS, Mem0, NeSTR, and TIDER, while TiMem and TsContriever still exhibit many retrieval misses, indicating that their retrievers often fail before temporal disambiguation can help. Overall, the diagnosis supports the paper's central claim: long-horizon memory QA is bottlenecked by retrieving the right evidence from the right time, not merely by answer-time reasoning.

\subsection{Case Studies}
\label{app:diagnostic:cases}

We summarize the recurring baseline failure modes observed across cases and then walk through three representative cases covering single-, dual-, and multi-window queries.

\paragraph{Common baseline failure modes.}
Baseline mistakes fall into five recurring patterns: (i) \emph{topic shell only}---retrieved context preserves the right person and broad theme but loses the specific event core (e.g., MemOS, Zep on \texttt{q100}, \texttt{q181}); (ii) \emph{event substitution with temporal drift}---a same-person event from a nearby month replaces the queried one (e.g., TiMem and Mem0 on \texttt{q100}, \texttt{q181}); (iii) \emph{long-term persona suppression}---high-frequency identity facts override the question's specific temporal window (e.g., MemOS and TsContriever on \texttt{q143}); (iv) \emph{single-endpoint capture}---one of two required temporal anchors is retrieved while the other is dropped (e.g., MemOS, NeSTR on \texttt{q120}, \texttt{q193}); and (v) \emph{near-total miss}---retrieval finds little relevant content (e.g., TsContriever on \texttt{q100}, \texttt{q164}). The three walk-throughs below illustrate (i)--(iv); failure mode (v) is largely confined to TsContriever and is consistent with its low overall accuracy in Table~\ref{tab:category-acc}.

\paragraph{Case 1: same-entity multi-event disambiguation (\texttt{user0/q100}).}
The question asks what trouble Huang Wenhui ran into in March 2025; the gold answer is that her pre-sold apartment project halted construction amid rumors that the developer's cash flow had broken down. Within the same conversational thread, Huang Wenhui appears in at least five negatively-toned events spread across the year---a January cafe scam (\texttt{q181}), the March apartment halt, an April game-project conflict (\texttt{q28}), April/July neighbor-dog disturbances (\texttt{q98}, \texttt{q37}), and a September app-launch slump (\texttt{q49})---so the challenge is isolating the March incident from this same-entity cluster. TIDER retrieves the March 12 messages that explicitly mention the halted construction site, the broken developer cash flow, and the continuing mortgage pressure. MemOS and Zep keep only the topic shell (``A-Hui has a housing problem''); TiMem replaces the event with a later apartment-water-leak story; Mem0 drifts to the July neighbor-dog incident; NeSTR preserves the negative-affect shell but slides to the January cafe scam. All baselines identify the correct entity, but only TIDER recovers the correct month-event pair.

\paragraph{Case 2: dual-window endpoint coverage (\texttt{user1/q120}).}
The question asks for the user's core activities on the first weekend after moving (June 14--15). The gold answer requires a complete two-day chain: a park walk and a housewarming dinner on day one, then a Pilates session and a pottery-class breakthrough on day two. TIDER retrieves evidence for all four components, with both days' messages co-occurring in its retrieved context, and answers the chain completely. MemOS retains the park-walk fragment but drops the second-day activities; TiMem and Zep pull in broadly related ``new-home'' or ``weekend'' memories without preserving the precise two-day structure; NeSTR captures the park and pottery pieces but misses Pilates. The challenge here is not topical relevance but preserving every answer-bearing endpoint required by a temporally compositional question.

\paragraph{Case 3: stage-level multi-window summarization (\texttt{user1/q143}).}
The question asks what the user's main work content was at Zhengzhou Central Hospital between August 11 and August 27; the gold answer is a respiratory-medicine rotation, including community-acquired pneumonia admissions, a COPD exacerbation case, a difficult-case discussion meeting, and the associated discharge documentation. TIDER's retrieved context anchors the August 11 admission and initial-course record, the August 21 case-discussion presentation, and the August 27 differential-diagnosis preparation, and additionally surfaces a ``reviewing the respiratory medicine section'' note. MemOS, Zep, and TsContriever all collapse back to the user's long-running cardiology persona (resident physician, cardiology rounds, cardiology case work), missing the temporary respiratory rotation entirely; TiMem fails to retrieve a coherent August 11--27 chain; Mem0 surfaces individual clinical actions but drops the rotation-stage label. This case illustrates how high-frequency persona evidence can override a specific time-bounded stage, and how TIDER's window-anchored retrieval resists this drift.

\section{Computational Resources}
\label{app:compute}

\subsection{Training Cost}
\label{app:compute:train}
We separate the SFT cold-start (Appendix~\ref{app:training:sft}) from the GRPO main loop (Appendix~\ref{app:training:grpo}). All training uses NVIDIA A100 80GB GPUs and the veRL framework. The SFT cold-start consumes about 3 wall-clock hours on $2\times$A100 80GB; a single GRPO run trains for $300$ steps in roughly 100 hours on $2\times$A100 80GB.
\subsection{Inference Latency}
\label{app:compute:latency}
We measure the marginal latency introduced by TIDER's temporal channel on the LifeBench-EN test split, holding the GPT-5-mini answer step fixed and toggling only the temporal pathway. The 1.7B locator is served on a single A100 80GB via vLLM. Averaged across $408$ questions, enabling the temporal channel adds $0.41$s of wall-clock latency per query relative to the no-temporal baseline; this overhead aggregates the locator forward pass, the timeline-summary retrieval branch, and the distance-decay rerank that fuses the two channels. The figure is well below the cost of the downstream answer call, which dominates end-to-end response time at the several-second scale, so the dual-channel design recovers the accuracy gap reported in Section~\ref{sec:experiments} without bottlenecking the GPT-5-mini~/ GPT-4o-mini answer step.

\section{Limitations and Broader Impact}
\label{app:impact}

\subsection{Limitations}
\label{app:impact:limits}
We discuss three limitations of TIDER that are visible from the current setup and shape the most natural next steps; each isolates a distinct failure axis (data realism, upstream memory dependency, language coverage) so that the relative weight of remediation effort is clear to the reader.

\paragraph{Synthetic personal histories.}
LifeBench is generated by an anonymized-persona-driven simulator with explicit temporal anchoring and cross-source consistency checks, and LoCoMo is multi-session synthetic dialogue rather than real long-term user data. Real personal traces likely contain heavier noise, more frequent timestamp gaps, and stronger cross-source contradictions than either benchmark exposes. Validating TIDER on a real, consented personal-memory corpus---under the privacy and retention safeguards discussed in Appendix~\ref{app:impact:broader}---therefore remains an open direction, and the gains we report should be read as upper bounds achievable when the temporal signal is well-formed.

\paragraph{Upstream memory-item dependency.}
TIDER's temporal channel operates on a query-adaptive timeline summary $\mathcal{S}_q$ drawn from clustered memory items in the memory pool $\mathcal{M}$ (Appendix~\ref{app:memory-pool}). We deliberately keep this construction as simple as possible---temporally bounded clustering followed by a single-pass summarization prompt---so that the locator does not couple to a bespoke memory representation; even so, $\mathcal{S}_q$ still inherits the quality of upstream item extraction, since mis-dated, mis-attributed, or fragmented items propagate into the cluster boundaries and into the resulting summary. The locator can only filter on this view, so its window predictions degrade whenever the upstream memory pool is noisy. 

\paragraph{Language scope.}
LifeBench and LoCoMo are both English-only, so TIDER's evaluation does not cover languages whose date conventions diverge from the Gregorian numerical form (e.g., Chinese lunar dates, Japanese era names) or whose temporal expressions carry richer morphological cues. The locator prompt and the gold-window construction both rely on the parser shipped with our extractor, which is calibrated for English; cross-lingual deployment will likely require multilingual date normalization upstream of the temporal channel before the locator's window predictions become meaningful.

\subsection{Broader Impact}
\label{app:impact:broader}
TIDER targets long-term personalized assistants whose factual grounding rests on a temporally anchored memory pool; we discuss its societal impact along three positive axes and two risk axes, and note one experimental safeguard. On the positive side: (i) the dual-channel design strengthens the reliability of long-term personal-assistant deployments that must reason over months or years of dialogue history; (ii) by filtering candidates against query-relevant temporal windows rather than reweighting them after generation, the temporal channel directly reduces cross-time evidence hallucinations at the retrieval stage; and (iii) the locator's window predictions $W_\text{pred}$ are user-inspectable, exposing which time region the system treated as relevant and giving end users an auditable handle on retrieval decisions. Two principal risks remain: (i) an adversary who can write into the dialogue stream may craft contradictory or fabricated dates that push the locator toward an incorrect window, thereby suppressing or surfacing specific memories at retrieval time; and (ii) any bias in the upstream LLM-based extractor---including event misattribution along demographic or linguistic lines---propagates directly into the temporally bounded clusters and the resulting timeline summary $\mathcal{S}_q$, so locator decisions inherit those biases without an independent corrective step. As an experimental safeguard, all training and evaluation in this paper use synthetic benchmarks (LifeBench, LoCoMo); no real consented personal-memory corpus was used at any stage.

\section{Prompt Templates and Judge Validation}
\label{app:prompts}

This appendix lists the prompts used by the memory pool builder, the locator, and the two LLM judges in the exact form fed to the underlying LLMs (\S\ref{app:prompts:cluster}--\S\ref{app:prompts:judge-evd}), and reports a cross-judge consistency audit (\S\ref{app:judge-consistency}).

\subsection{Memory Pool Construction Prompts}
\label{app:prompts:cluster}
The two prompts that drive memory-pool construction are listed below.

\paragraph{Cluster summarization (\texttt{REORGANIZE\_PROMPT}).} Used to merge the items in a temporally bounded cluster (or sub-cluster) into a single chronologically-ordered event summary, with date extraction and entity disambiguation handled inline.

\begin{promptbox}[title=Cluster summarization prompt, colbacktitle=SteelBlue!90!black, colframe=SteelBlue!90!black]
You are a memory clustering and summarization expert.

Given the following child memory items:

{memory_items_text}

Please perform:
1. Identify information that reflects user's experiences, beliefs,
concerns, decisions, plans, or reactions -- including meaningful
input from assistant that user acknowledged or responded to.
2. Resolve all time, person, and event references clearly:
- Convert relative time expressions (e.g., "yesterday," "next Friday")
into absolute dates using the message timestamp if possible.
- Clearly distinguish between event time and message time.
- If uncertainty exists, state it explicitly (e.g., "around June
2025," "exact date unclear").
- Include specific locations if mentioned.
- Resolve all pronouns, aliases, and ambiguous references into full
names or identities.
- Disambiguate people with the same name if applicable.
3. This summary is a navigation aid for retrieval, not canonical
evidence. You may use third-person for clarity, but do not rewrite
into a polished narrative. Do not collapse distinct event instances
across different dates into one fact. Do not introduce any detail
not present in the child items.
4. Do not omit any information that user is likely to remember.
- Include all key experiences, thoughts, emotional responses, and
plans -- even if they seem minor.
- Prioritize completeness and fidelity over conciseness.
- Do not generalize or skip details that could be personally
meaningful to user.
5. Summarize all child memory items into one memory item.

Output format requirements:
- The value field must be a natural language paragraph, describing
events in chronological order.
- Absolutely forbidden to output JSON arrays, lists, or structured data
formats (e.g., [{...}, {...}]).
- Each independent event starts with "YYYY-MM-DD," followed by a
sentence. Sentences are separated by periods.

Language rules:
- The `key`, `value`, `tags`, `summary` fields must match the mostly
used language of the input memory items.
- Keep `memory_type` in English.

Return valid JSON:
{
    "key": <string, a concise title of the `value` field>,
    "memory_type": <string, "LongTermMemory" or "UserMemory">,
    "value": <natural language paragraph describing events in
    chronological order; preserve dates, amounts, names,
    locations; do not collapse different event instances into
    one abstract statement; do not introduce facts not present
    in the child items>,
    "tags": <list of relevant thematic keywords>,
    "summary": <natural paragraph from user's perspective, 120-200 words,
    same language as the input>
}
\end{promptbox}

\paragraph{Sub-cluster split.} Used to subdivide an oversized cluster into 2--10-item sub-clusters before summarization.

\begin{promptbox}[title=Sub-cluster split prompt, colbacktitle=SeaGreen!90!black, colframe=SeaGreen!90!black]
You are a memory organization expert.

You are given a cluster of memory items, each with an ID and content.
Your task is to divide these into smaller, semantically meaningful
sub-clusters.

Instructions:
- Identify natural topics by analyzing common time, place, people, and
event elements.
- Each sub-cluster must reflect a coherent theme that helps retrieval.
- Each sub-cluster should have 2-10 items. Discard singletons.
- Each item ID must appear in exactly one sub-cluster or be discarded.
No duplicates are allowed.
- All IDs in the output must be from the provided Memory items.
- Return strictly valid JSON only.

Example: If you have items about a project across multiple phases,
group them by milestone, team, or event.

Language rules:
- The `key` fields must match the mostly used language of the
clustered memories.

Return valid JSON:
{
    "clusters": [
    {"ids": ["<id1>", "<id2>", ...], "key": "<concise memory title>"},
    ...
    ]
}

Memory items:
{joined_scene}
\end{promptbox}

\subsection{Locator Prompt and Output Schema}
\label{app:prompts:locator}
The locator $\pi_\theta$ emits a single JSON object with two fields per query: a temporal-sensitivity flag and a list of temporal windows.

\begin{promptbox}[title=Locator output schema, colbacktitle=RoyalBlue!90!black, colframe=RoyalBlue!90!black]
{
    "detect": true | false,
    "windows": [
    {"start": "YYYY-MM-DD", "end": "YYYY-MM-DD"}, ...
    ]
}
\end{promptbox}

The \texttt{detect} field follows the temporal-sensitivity definition listed below, which is the prompt fed to GPT-5.2 to label \texttt{gold\_temporal} during training-data construction (Appendix~\ref{app:training}); $\pi_\theta$ is trained to imitate this judgment. The prompt text itself uses the \texttt{detect} key so that the annotation format matches the locator schema. Window prediction is supervised directly from $(q, W_\text{gold})$ pairs and does not require a separate LLM prompt template at data-construction time.

\begin{promptbox}[title=Temporal-sensitivity detection prompt, colbacktitle=RoyalBlue!90!black, colframe=RoyalBlue!90!black]
You are a temporal constraint detector. Determine whether the user's
question requires searching memories within a specific time range.

## What is a temporal constraint
A temporal constraint limits the time range for memory retrieval. Two
types:
- Explicit: directly mentions time, e.g. "in June 2025", "last month",
"during Spring Festival", "in the last 3 days"
- Implicit: indirectly limits time via events, e.g. "after mom got
sick", "before moving", "after that business trip"
- Event-pair (time gap): asks how long between two events -- both
events are implicit constraints; extract the events, NOT the time-gap
word ("how long", "how many days")

## What is NOT a temporal constraint (important)
- Factual questions: "When is the birthday", "What time do you wake up"
-- asking for information, not limiting search range
- Pure semantic questions: "favorite restaurant", "how do you commute"
- Dates in multiple-choice options (e.g. "A. Jan 6  B. Jan 8") are
answer choices, not search constraints
- Time words that don't change meaning if removed: "the restaurant
mentioned before" -- "before" is casual speech, not a constraint
- Time-gap words are NOT expressions: "how many days", "how long",
"how far in advance" are the answer target -- do NOT put them in
expressions

## Expression extraction rules
1. Every temporal expression must be extracted separately -- do not
omit any.
2. Explicit dates (e.g. "June 2025", "April 26", "Spring Festival")
must never be omitted, even when other expressions like "recently"
are also present.
3. Do NOT merge time words with event content -- "recently changed
jobs" yields "recently", not "recently changed jobs".

## Decision method
Ask yourself: does answering this question require limiting the search
to a specific time period? If not, it's false.

## User question
${query}

Output strict JSON only (no other text):
{
    "detect": true/false,
    "expressions": []
}
Fill expressions with the original temporal expressions extracted from
the query. Empty array if no constraint.

## Examples
### Positive (explicit)
Input: In May 2025, what did Zhang Ming's cat eat?
Output: {"detect": true,
    "expressions": ["In May 2025"]}

### Positive (multiple temporal expressions)
Input: In June 2025, Zhang Ming mentioned on the phone that he
recently changed jobs. What was the new job?
Output: {"detect": true,
    "expressions": ["In June 2025", "recently"]}

### Positive (implicit)
Input: What pets did he get after moving?
Output: {"detect": true,
    "expressions": ["after moving"]}

### Negative (factual question)
Input: What is Zhang Ming's favorite restaurant?
Output: {"detect": false, "expressions": []}

### Negative (dates in options)
Input: When did Zhang Ming go to the hospital? A. Jan 6 B. Jan 8
C. Jan 10
Output: {"detect": false, "expressions": []}

### Positive (event anchors, not time words)
Input: How many days before checking in did she book the guesthouse?
Output: {"detect": true,
    "expressions": ["book the guesthouse", "checking in"]}

### Positive (event-pair time gap)
Input: How many days passed between Zhang Ming signing the lease and
actually moving in?
Output: 
{
"detect": true,
"expressions": ["signing the lease", "actually moving in"]
}
\end{promptbox}

%
%
%
%
%

\subsection{LLM Judge for Answer Correctness ($r_\text{out}$)}
\label{app:prompts:judge-out}
Answer generation and answer judging are kept separate. The downstream answer is generated by a frozen LLM, but the answer-generation prompt is dataset-specific: following the LifeBench paper's official evaluation requirement, LifeBench uses the official \texttt{QA\_PROMPT} from the LoCoMo reference implementation, whereas LoCoMo uses the Mem0-style \texttt{ANSWER\_PROMPT\_MEM0}. In both datasets, answer correctness is judged with the official LoCoMo prompt reproduced below; we invoke the same judge prompt independently three times and use majority vote (CORRECT in at least two of three calls) as the final label.

\paragraph{LifeBench answer generation.}
For LifeBench, the retrieved context is prepended before this prompt suffix.
\begin{promptbox}[title=LifeBench answer generation prompt, colbacktitle=DarkOrange!90!black, colframe=DarkOrange!90!black]
Based on the above context, write an answer in the form of a short
phrase for the following question. Answer with exact words from the
context whenever possible.

Question: {} Short answer:
\end{promptbox}

\paragraph{LoCoMo answer generation.}
The following Mem0-style answer prompt is used for LoCoMo, not for LifeBench.
\begin{promptbox}[title=LoCoMo answer generation prompt, colbacktitle=DarkOrange!90!black, colframe=DarkOrange!90!black]
You are an intelligent memory assistant tasked with retrieving accurate
information from conversation memories.

# CONTEXT:
You have access to memories from two speakers in a conversation. These
memories contain timestamped information that may be relevant to
answering the question.

# INSTRUCTIONS:
1. Carefully analyze all provided memories from both speakers
2. Pay special attention to the timestamps to determine the answer
3. If the question asks about a specific event or fact, look for
direct evidence in the memories
4. If the memories contain contradictory information, prioritize the
most recent memory
5. If there is a question about time references (like "last year",
"two months ago", etc.), calculate the actual date based on the
memory timestamp. For example, if a memory from 4 May 2022 mentions
"went to India last year," then the trip occurred in 2021.
6. Always convert relative time references to specific dates, months,
or years. For example, convert "last year" to "2022" or "two months
ago" to "March 2023" based on the memory timestamp. Ignore the
reference while answering the question.
7. Focus only on the content of the memories from both speakers. Do
not confuse character names mentioned in memories with the actual
users who created those memories.
8. The answer should be less than 5-6 words.

# APPROACH (Think step by step):
1. First, examine all memories that contain information related to
the question
2. Examine the timestamps and content of these memories carefully
3. Look for explicit mentions of dates, times, locations, or events
that answer the question
4. If the answer requires calculation (e.g., converting relative time
references), show your work
5. Formulate a precise, concise answer based solely on the evidence
in the memories
6. Double-check that your answer directly addresses the question
asked
7. Ensure your final answer is specific and avoids vague time
references

{context}

Question: {question}

Answer:
\end{promptbox}

\paragraph{Answer-correctness judge.}
\begin{promptbox}[title=Answer-correctness judge prompt, colbacktitle=FireBrick!90!black, colframe=FireBrick!90!black]
[system]
You are an expert grader that determines if answers to questions match
a gold standard answer

[user]
Your task is to label an answer to a question as 'CORRECT' or 'WRONG'.
You will be given the following data:
(1) a question (posed by one user to another user),
(2) a 'gold' (ground truth) answer,
(3) a generated answer
which you will score as CORRECT/WRONG.

The point of the question is to ask about something one user should
know about the other user based on their prior conversations.
The gold answer will usually be a concise and short answer that
includes the referenced topic, for example:
Question: Do you remember what I got the last time I went to Hawaii?
Gold answer: A shell necklace
The generated answer might be much longer, but you should be generous
with your grading - as long as it touches on the same topic as the
gold answer, it should be counted as CORRECT.

For time related questions, the gold answer will be a specific date,
month, year, etc. The generated answer might be much longer or use
relative time references (like "last Tuesday" or "next month"), but
you should be generous with your grading - as long as it refers to
the same date or time period as the gold answer, it should be counted
as CORRECT. Even if the format differs (e.g., "May 7th" vs "7 May"),
consider it CORRECT if it's the same date.

Now it's time for the real question:
Question: {question}
Gold answer: {gold_answer}
Generated answer: {response}

First, provide a short (one sentence) explanation of your reasoning,
then finish with CORRECT or WRONG.
Do NOT include both CORRECT and WRONG in your response, or it will
break the evaluation script.

Just return the label CORRECT or WRONG in a json format with the key
as "label".
\end{promptbox}

\subsection{LLM Judge for Evidence Coverage ($r_\text{evd}$)}
\label{app:prompts:judge-evd}
The evidence-coverage judge $r_\text{evd}$ is a reward that measures item-level evidence recall: for each question, the reviewer independently checks every gold evidence item and returns $r_\text{evd}=m/n$, where $n$ is the number of gold evidence items and $m$ is the number of items explicitly covered by the retrieved memory set $\hat{\mathcal{E}}$. The reward is therefore a floating-point value in $[0,1]$.

\paragraph{Evidence-coverage judge.}
\begin{promptbox}[title=Evidence-coverage judge prompt, colbacktitle=FireBrick!90!black, colframe=FireBrick!90!black]
[system]
You are a strict evidence-coverage reviewer.

Given a list of gold evidence facts and retrieved memories, decide
what fraction of gold evidence facts are covered by the retrieved memories.

Coverage criteria (strict):
- Each gold evidence fact is judged independently.
- A gold fact is covered only if at least one retrieved memory
explicitly mentions semantically identical content.
- The retrieved text must mention the gold fact's specific person
names, places, events, numbers, dates, and key modifiers.
- Conceptually related text, partial hits, loose paraphrases, and
superordinate substitutions do NOT count as coverage.
- Synonyms are allowed.
For example, "xiao-long-bao" = "steamed soup dumplings".
- Category broadening is not allowed.
For example, "xiao-long-bao" != "dim sum".
- Missing a key modifier counts as not covered.
For example, "Li Ming ate xiao-long-bao" !=
"Li Ming ate something".

Compute:

coverage = number of covered gold facts / total number of gold facts

Return only JSON in this format:

{
    "coverage": <float between 0.0 and 1.0>,
    "reason": "<one short English sentence>"
}

Positive 1:
gold:
["Li Ming ate xiao-long-bao in Lujiazui, Shanghai"]

retrieved:
["I went with Li Ming to try xiao-long-bao in Lujiazui,
Shanghai -- it was delicious"]

output:
{
    "coverage": 1.0,
    "reason": "the food and location are identical"
}

Negative 1, superordinate substitution:
gold:
["Li Ming ate xiao-long-bao in Lujiazui, Shanghai"]

retrieved:
["Li Ming ate dim sum in Shanghai"]

output:
{
    "coverage": 0.0,
    "reason": "dim sum is not equal to xiao-long-bao"
}

Negative 2, partial coverage:
gold:
[
"Li Ming went to Shanghai on April 1",
"Li Ming ate xiao-long-bao in Shanghai"
]

retrieved:
["Li Ming went to Shanghai on April 1 for a meeting"]

output:
{
    "coverage": 0.5,
    "reason": "only the Shanghai trip on April 1 is covered"
}

Negative 3, number mismatch:
gold:
["Li Ming spent 200 yuan on a book"]

retrieved:
["Li Ming bought a book"]

output:
{
    "coverage": 0.0,
    "reason": "the price is missing"
}

[user]
gold_evidence:
{gold}

retrieved_memories:
{retrieved}

Output JSON following the rules above.
\end{promptbox}


\subsection{Error-Cause Judge Prompt}
\label{app:prompts:error-cause}
The following prompt is used for the LLM reviewer in Appendix~\ref{app:diagnostic:error-cause}.

\begin{promptbox}[title=Error-cause judge prompt, colbacktitle=FireBrick!90!black, colframe=FireBrick!90!black]
[system]
You are a strict retrieval failure diagnosis reviewer for
long-horizon personal-memory QA.

Role:
You diagnose why an incorrect answer happened. Your main focus is
retrieval quality, especially whether retrieval selected evidence from
the wrong time or wrong event instance.

Task:
Given one incorrect QA case, compare:
1) question,
2) gold answer,
3) model answer,
4) gold evidence,
5) retrieved context.
Assign exactly one primary_error.

Allowed labels:

1. temporal_mismatch
Use when the retrieved context contains a salient distractor that is
semantically related to the question but temporally incompatible with
the gold evidence. This requires at least one of:
- same person/entity/topic, but a clearly different date/window/event
instance than the gold evidence;
- stale state or old preference used when the question requires a later
updated state;
- future plan/reminder/preview used instead of the actual completed
event;
- long-term persona or routine activity replacing a temporary
time-bounded stage;
- model answer is visibly based on a retrieved wrong-time event.

2. partial_evidence
Use when retrieved context contains correct evidence from the right
event/window but misses other necessary correct evidence. Typical cases:
one hop of a multi-hop query, one endpoint of a dual-window query, one
item in a list, or correct event but missing a key detail. 

3. retrieval_miss
Use when the answer-bearing evidence is absent and there is no explicit
wrong-time same-entity/same-topic distractor. This includes off-topic
retrieval, generic persona summaries, vague related memories without a
date/event-instance conflict, and near-empty contexts.

4. reason_failure
Use when retrieved context contains enough correct evidence from the
right event/window to answer correctly, but the model answer is wrong
due to reasoning, aggregation, option selection, or wording.

5. other
Use only when none of the above fits.

Decision procedure:
Step 1: Is there enough correct retrieved evidence from the right
event/window for a careful reader to answer? If yes and answer is still
wrong, choose reason_failure.
Step 2: Is some correct right-event/right-window evidence present but
incomplete? If yes, choose partial_evidence unless a wrong-time
distractor is clearly dominant.
Step 3: Is there an explicit wrong-time/stale/future/wrong-stage
distractor related to the same entity/topic and likely driving the wrong
answer? If yes, choose temporal_mismatch.
Step 4: Otherwise choose retrieval_miss.

Important constraints:
- Do NOT label temporal_mismatch just because the benchmark is temporal.
- Do NOT label temporal_mismatch for non-temporal questions unless the
failure truly depends on stale/updated state or a wrong event instance.
- Do NOT label temporal_mismatch when retrieved context is merely
broad/generic and lacks a clear wrong date/window/event.
- The explanation for temporal_mismatch must name the wrong-time
distractor type, e.g., "July dog incident replaces March apartment
halt" or "cardiology persona replaces August respiratory rotation".

Example A: temporal_mismatch
question: "What trouble did Huang Wenhui face in March?"
gold evidence: "2025-03-12: her pre-sold apartment project halted
construction and the developer's cash flow broke down."
retrieved context: "2025-07-10: Huang Wenhui complained about a
neighbor's dog disturbing her sleep."
model answer: "A neighbor's dog bothered her."
output: {"primary_error":"temporal_mismatch",
    "secondary_errors":["retrieval_miss"],
    "retrieval_contains_core_evidence":false,
    "retrieval_contains_wrong_time_related_evidence":true,
    "explanation":"the July dog incident replaces the queried March
    apartment halt"}

Example B: retrieval_miss, not temporal_mismatch
question: "What trouble did Huang Wenhui face in March?"
gold evidence: "2025-03-12: her pre-sold apartment project halted
construction."
retrieved context: "Huang Wenhui has been stressed about housing and
work."
model answer: "She had general housing stress."
output: {"primary_error":"retrieval_miss","secondary_errors":[],
    "retrieval_contains_core_evidence":false,
    "retrieval_contains_wrong_time_related_evidence":false,
    "explanation":"the context is generic and lacks a specific wrong-time
    event"}

Example C: partial_evidence
question: "What did she do on both days of the first weekend after
moving?"
gold evidence: ["Saturday: park walk and housewarming dinner",
"Sunday: Pilates and pottery class"]
retrieved context: "Saturday: she took a park walk and had a
housewarming dinner."
model answer: "She walked in the park and had dinner."
output: {"primary_error":"partial_evidence","secondary_errors":[],
    "retrieval_contains_core_evidence":true,
    "retrieval_contains_wrong_time_related_evidence":false,
    "explanation":"the correct weekend is retrieved but only the Saturday
    endpoint is covered"}

Example D: reason_failure
question: "Which option is the appointment date?"
gold evidence: "The app notification confirmed January 10."
retrieved context: "The app notification confirmed the appointment on
January 10."
model answer: "B: January 8."
output: {"primary_error":"reason_failure","secondary_errors":[],
    "retrieval_contains_core_evidence":true,
    "retrieval_contains_wrong_time_related_evidence":false,
    "explanation":"the right evidence is present but the option selection
    is wrong"}

Return only valid JSON:
{
    "primary_error": "temporal_mismatch | partial_evidence | retrieval_miss | reason_failure | other",
    "secondary_errors": ["temporal_mismatch | partial_evidence | retrieval_miss | reason_failure | other"],
    "retrieval_contains_core_evidence": true/false,
    "retrieval_contains_wrong_time_related_evidence": true/false,
    "explanation": "one concise English sentence"
}

[user]
question:
{question}

gold_answer:
{gold_answer}

model_answer:
{model_answer}

gold_evidence:
{gold_evidence}

retrieved_context:
{retrieved_context}
\end{promptbox}

\subsection{Judge Consistency Audit}
\label{app:judge-consistency}

This appendix audits the variance of the LLM judges used for GRPO.

To audit cross-model agreement, we re-evaluate the same response set ($n=408$ from LifeBench user-0 and user-1) with two additional judges drawn from different model families---\texttt{claude-sonnet-4-6} (Anthropic) and \texttt{gemini-3.1-flash-lite-preview-nothink} (Google)---using the exact same system and user prompts as the \texttt{gpt-5-mini} judge reported in the main paper.
%

%

\begin{center}
	\begin{tabular}{lccc}
		\toprule
		& \texttt{gpt-5-mini} & \texttt{claude-sonnet-4-6} & \texttt{gemini-3.1-lite} \\
		\midrule
		\texttt{gpt-5-mini} & --- & 0.894 & 0.836 \\
		\texttt{claude-sonnet-4-6} & 0.894 & --- & 0.894 \\
		\texttt{gemini-3.1-lite} & 0.836 & 0.894 & --- \\
		\bottomrule
	\end{tabular}
\end{center}

All three pairwise Cohen's $\kappa$ values lie in $[0.836, 0.894]$, well above the $0.81$ threshold conventionally interpreted as ``almost perfect'' agreement.
We therefore report a single judge (\texttt{gpt-5-mini}) throughout the main paper without loss of generality.
%

%% file: checklist.tex
\section*{NeurIPS Paper Checklist}

\begin{enumerate}
	
	\item {\bf Claims}
	\item[] Question: Do the main claims made in the abstract and introduction accurately reflect the paper's contributions and scope?
	\item[] Answer: \answerYes{} 
	\item[] Justification: The abstract and introduction state the TIDER framework, the RL-trained temporal interval locator, the dual-channel retrieval mechanism, and the reported gains on LifeBench and LoCoMo. These claims are supported by the main results in Section~\ref{sec:results} and ablations in Section~\ref{sec:ablation}.
	
	\item {\bf Limitations}
	\item[] Question: Does the paper discuss the limitations of the work performed by the authors?
	\item[] Answer: \answerYes{} 
	\item[] Justification: Appendix~\ref{app:impact:limits} discusses limitations from synthetic personal histories, dependence on upstream memory-item extraction, and English-only evaluation.
	
	\item {\bf Theory assumptions and proofs}
	\item[] Question: For each theoretical result, does the paper provide the full set of assumptions and a complete (and correct) proof?
	\item[] Answer: \answerNA{} 
	\item[] Justification: The paper proposes an empirical retrieval framework and does not claim theoretical results requiring formal proofs.
	
	\item {\bf Experimental result reproducibility}
	\item[] Question: Does the paper fully disclose all the information needed to reproduce the main experimental results of the paper to the extent that it affects the main claims and/or conclusions of the paper (regardless of whether the code and data are provided or not)?
	\item[] Answer: \answerYes{} 
	\item[] Justification: Sections~\ref{sec:datasets}--\ref{sec:setup} and Appendices~\ref{app:datasets}, \ref{app:memory-pool}, \ref{app:training}, and \ref{app:prompts} describe the datasets, splits, memory-pool construction, training procedure, hyperparameters, prompts, and evaluation protocol needed to reproduce the main results. The supplementary Markdown guide and anonymized code package provide executable scripts and configuration files for reproducing the main results.

	\item {\bf Open access to data and code}
	\item[] Question: Does the paper provide open access to the data and code, with sufficient instructions to faithfully reproduce the main experimental results, as described in supplemental material?
	\item[] Answer: \answerYes{} 
	\item[] Justification: We provide an anonymized supplementary code package with a Markdown reproduction guide, including environment setup, data preparation instructions, configuration files, and scripts for reproducing the main experiments.

	\item {\bf Experimental setting/details}
	\item[] Question: Does the paper specify all the training and test details (e.g., data splits, hyperparameters, how they were chosen, type of optimizer) necessary to understand the results?
	\item[] Answer: \answerYes{} 
	\item[] Justification: Section~\ref{sec:setup} specifies datasets, baselines, retrieval budget, model choices, split construction, and key hyperparameters; Appendix~\ref{app:training} gives SFT and GRPO training details, and Appendix~\ref{app:baselines} details baseline adaptation. The supplementary code package includes the corresponding configuration files and run scripts.
	
	\item {\bf Experiment statistical significance}
	\item[] Question: Does the paper report error bars suitably and correctly defined or other appropriate information about the statistical significance of the experiments?
	\item[] Answer: \answerYes{} 
	\item[] Justification: The main result tables report means and standard deviations over independent evaluation runs, and Appendix~\ref{app:quantitative} reports stratified accuracy with standard deviations.
	
	\item {\bf Experiments compute resources}
	\item[] Question: For each experiment, does the paper provide sufficient information on the computer resources (type of compute workers, memory, time of execution) needed to reproduce the experiments?
	\item[] Answer: \answerYes{} 
	\item[] Justification: Appendix~\ref{app:compute} reports training hardware, wall-clock cost, and the per-query inference latency.
	
	\item {\bf Code of ethics}
	\item[] Question: Does the research conducted in the paper conform, in every respect, with the NeurIPS Code of Ethics \url{https://neurips.cc/public/EthicsGuidelines}?
	\item[] Answer: \answerYes{} 
	\item[] Justification: The work uses synthetic benchmarks and does not use real personal-memory data; privacy and misuse risks are discussed in Appendix~\ref{app:impact:broader}. The submission is anonymized and does not disclose identifying author information.

	\item {\bf Broader impacts}
	\item[] Question: Does the paper discuss both potential positive societal impacts and negative societal impacts of the work performed?
	\item[] Answer: \answerYes{} 
	\item[] Justification: Appendix~\ref{app:impact:broader} discusses positive impacts for long-term personalized assistants as well as risks from fabricated dates, biased extraction, and propagation of upstream memory errors.
	
	\item {\bf Safeguards}
	\item[] Question: Does the paper describe safeguards that have been put in place for responsible release of data or models that have a high risk for misuse (e.g., pre-trained language models, image generators, or scraped datasets)?
	\item[] Answer: \answerNA{} 
	\item[] Justification: The paper does not release a high-risk foundation model, scraped dataset, image generator, or real personal-memory corpus. As an experimental safeguard, all training and evaluation use synthetic benchmarks.
	
	\item {\bf Licenses for existing assets}
	\item[] Question: Are the creators or original owners of assets (e.g., code, data, models), used in the paper, properly credited and are the license and terms of use explicitly mentioned and properly respected?
	\item[] Answer: \answerYes{} 
	\item[] Justification: Existing assets used in the paper, including LifeBench, LoCoMo, Qwen3-1.7B, BGE-M3, BM25, veRL, and API-based LLMs, are credited through citations in the paper. The supplementary Markdown guide lists their project links, versions when available, and license or terms-of-use information, and our use follows those terms.
	
	\item {\bf New assets}
	\item[] Question: Are new assets introduced in the paper well documented and is the documentation provided alongside the assets?
	\item[] Answer: \answerYes{} 
	\item[] Justification: We introduce an anonymized supplementary code package as a new asset. It is documented with a Markdown guide describing its purpose, setup, reproduction commands, dependencies, and license.
	
	\item {\bf Crowdsourcing and research with human subjects}
	\item[] Question: For crowdsourcing experiments and research with human subjects, does the paper include the full text of instructions given to participants and screenshots, if applicable, as well as details about compensation (if any)? 
	\item[] Answer: \answerNA{} 
	\item[] Justification: The work does not conduct crowdsourcing experiments or human-subject studies. The only human involvement described is verification of synthetic benchmark-derived temporal labels, not collection of human-subject data.
	
	\item {\bf Institutional review board (IRB) approvals or equivalent for research with human subjects}
	\item[] Question: Does the paper describe potential risks incurred by study participants, whether such risks were disclosed to the subjects, and whether Institutional Review Board (IRB) approvals (or an equivalent approval/review based on the requirements of your country or institution) were obtained?
	\item[] Answer: \answerNA{} 
	\item[] Justification: The paper does not involve human-subject research or real personal-memory data, so IRB approval is not applicable.
	
	\item {\bf Declaration of LLM usage}
	\item[] Question: Does the paper describe the usage of LLMs if it is an important, original, or non-standard component of the core methods in this research? Note that if the LLM is used only for writing, editing, or formatting purposes and does \emph{not} impact the core methodology, scientific rigor, or originality of the research, declaration is not required.
	\item[] Answer: \answerYes{} 
	\item[] Justification: LLMs are core components of the method and evaluation: they are used for memory-item extraction and summarization, temporal-label construction, downstream answer generation, and LLM-based judging. These uses are described in Sections~\ref{sec:method}--\ref{sec:setup} and Appendix~\ref{app:prompts}.
	
\end{enumerate}

%% file: neurips_2026.bbl
\begin{thebibliography}{35}
\providecommand{\natexlab}[1]{#1}
\providecommand{\url}[1]{\texttt{#1}}
\expandafter\ifx\csname urlstyle\endcsname\relax
  \providecommand{\doi}[1]{doi: #1}\else
  \providecommand{\doi}{doi: \begingroup \urlstyle{rm}\Url}\fi

\bibitem[Abdallah et~al.(2025)Abdallah, Piryani, Wallat, Anand, and Jatowt]{abdallah2025tempretrieverfusionbasedtemporaldense}
Abdelrahman Abdallah, Bhawna Piryani, Jonas Wallat, Avishek Anand, and Adam Jatowt.
\newblock Tempretriever: Fusion-based temporal dense passage retrieval for time-sensitive questions, 2025.
\newblock URL \url{https://arxiv.org/abs/2502.21024}.

\bibitem[Chen et~al.(2024{\natexlab{a}})Chen, Wang, Xu, Yuan, Zhang, Shi, Xie, Li, Yang, Zhu, Chen, Li, Chen, Hu, Wu, Ren, Fu, and Xiao]{chen2024personapersonalizationsurveyroleplaying}
Jiangjie Chen, Xintao Wang, Rui Xu, Siyu Yuan, Yikai Zhang, Wei Shi, Jian Xie, Shuang Li, Ruihan Yang, Tinghui Zhu, Aili Chen, Nianqi Li, Lida Chen, Caiyu Hu, Siye Wu, Scott Ren, Ziquan Fu, and Yanghua Xiao.
\newblock From persona to personalization: A survey on role-playing language agents, 2024{\natexlab{a}}.
\newblock URL \url{https://arxiv.org/abs/2404.18231}.

\bibitem[Chen et~al.(2024{\natexlab{b}})Chen, Xiao, Zhang, Luo, Lian, and Liu]{chen2024bgem3}
Jianlv Chen, Shitao Xiao, Peitian Zhang, Kun Luo, Defu Lian, and Zheng Liu.
\newblock {M3-Embedding}: Multi-linguality, multi-functionality, multi-granularity text embeddings through self-knowledge distillation, 2024{\natexlab{b}}.
\newblock URL \url{https://arxiv.org/abs/2402.03216}.

\bibitem[Cheng et~al.(2026)Cheng, Wang, Zhao, Ren, Chen, Xu, Huang, Chen, Li, Wang, Xie, Zhu, Jiang, Lu, Li, Wang, Liu, and Nguyen]{cheng2026lifebenchbenchmarklonghorizonmultisource}
Zihao Cheng, Weixin Wang, Yu~Zhao, Ziyang Ren, Jiaxuan Chen, Ruiyang Xu, Shuai Huang, Yang Chen, Guowei Li, Mengshi Wang, Yi~Xie, Ren Zhu, Zeren Jiang, Keda Lu, Yihong Li, Xiaoliang Wang, Liwei Liu, and Cam-Tu Nguyen.
\newblock Lifebench: A benchmark for long-horizon multi-source memory, 2026.
\newblock URL \url{https://arxiv.org/abs/2603.03781}.

\bibitem[Chhikara et~al.(2025)Chhikara, Khant, Aryan, Singh, and Yadav]{chhikara2025mem0buildingproductionreadyai}
Prateek Chhikara, Dev Khant, Saket Aryan, Taranjeet Singh, and Deshraj Yadav.
\newblock Mem0: Building production-ready ai agents with scalable long-term memory, 2025.
\newblock URL \url{https://arxiv.org/abs/2504.19413}.

\bibitem[Du et~al.(2026)Du, Wang, Xiang, Wang, Huang, XUE, Liang, Zeng, Mi, Bai, Shang, Pan, Jiang, and Wong]{du2026memoryt}
Yiming Du, Baojun Wang, Yifan Xiang, Zhaowei Wang, Wenyu Huang, Boyang XUE, Bin Liang, Xingshan Zeng, Fei Mi, Haoli Bai, Lifeng Shang, Jeff~Z. Pan, Yuxin Jiang, and Kam-Fai Wong.
\newblock Memory-t1: Reinforcement learning for temporal reasoning in multi-session agents.
\newblock In \emph{The Fourteenth International Conference on Learning Representations}, 2026.
\newblock URL \url{https://openreview.net/forum?id=vQf2YR2Kpd}.

\bibitem[Ge et~al.(2025)Ge, Romeo, Cai, Shu, Benajiba, Sunkara, and Zhang]{ge-etal-2025-tremu}
Yubin Ge, Salvatore Romeo, Jason Cai, Raphael Shu, Yassine Benajiba, Monica Sunkara, and Yi~Zhang.
\newblock {TR}e{M}u: Towards neuro-symbolic temporal reasoning for {LLM}-agents with memory in multi-session dialogues.
\newblock In Wanxiang Che, Joyce Nabende, Ekaterina Shutova, and Mohammad~Taher Pilehvar, editors, \emph{Findings of the Association for Computational Linguistics: ACL 2025}, pages 18974--18988, Vienna, Austria, July 2025. Association for Computational Linguistics.
\newblock ISBN 979-8-89176-256-5.
\newblock \doi{10.18653/v1/2025.findings-acl.972}.
\newblock URL \url{https://aclanthology.org/2025.findings-acl.972/}.

\bibitem[Han et~al.(2025)Han, Hwang, Cho, Jeong, Song, Lee, and Park]{han-etal-2025-temporal}
SeungYoon Han, Taeho Hwang, Sukmin Cho, Soyeong Jeong, Hoyun Song, Huije Lee, and Jong~C. Park.
\newblock Temporal information retrieval via time-specifier model merging.
\newblock In Yuji Zhang, Canyu Chen, Sha Li, Mor Geva, Chi Han, Xiaozhi Wang, Shangbin Feng, Silin Gao, Isabelle Augenstein, Mohit Bansal, Manling Li, and Heng Ji, editors, \emph{Proceedings of the 3rd Workshop on Towards Knowledgeable Foundation Models (KnowFM)}, pages 1--13, Vienna, Austria, August 2025. Association for Computational Linguistics.
\newblock ISBN 979-8-89176-283-1.
\newblock \doi{10.18653/v1/2025.knowllm-1.1}.
\newblock URL \url{https://aclanthology.org/2025.knowllm-1.1/}.

\bibitem[Healey et~al.(2019)Healey, Long, and Kahana]{healey2019contiguity}
M~Karl Healey, Nicole~M Long, and Michael~J Kahana.
\newblock Contiguity in episodic memory.
\newblock \emph{Psychonomic bulletin \& review}, 26\penalty0 (3):\penalty0 699--720, 2019.

\bibitem[Li et~al.(2025{\natexlab{a}})Li, Yang, Zhang, Deng, Wang, and Chua]{li2025helloagainllmpoweredpersonalized}
Hao Li, Chenghao Yang, An~Zhang, Yang Deng, Xiang Wang, and Tat-Seng Chua.
\newblock Hello again! llm-powered personalized agent for long-term dialogue, 2025{\natexlab{a}}.
\newblock URL \url{https://arxiv.org/abs/2406.05925}.

\bibitem[Li et~al.(2026)Li, Yu, Ni, Zeng, Xu, Zhang, Li, Sang, Duan, Wang, Liu, and Tan]{li2026timemtemporalhierarchicalmemoryconsolidation}
Kai Li, Xuanqing Yu, Ziyi Ni, Yi~Zeng, Yao Xu, Zheqing Zhang, Xin Li, Jitao Sang, Xiaogang Duan, Xuelei Wang, Chengbao Liu, and Jie Tan.
\newblock Timem: Temporal-hierarchical memory consolidation for long-horizon conversational agents, 2026.
\newblock URL \url{https://arxiv.org/abs/2601.02845}.

\bibitem[Li et~al.(2025{\natexlab{b}})Li, Song, Wang, Niu, Chen, Yang, Xi, Lai, Zhao, Wang, Ren, Lin, Huo, Chen, Chen, Li, Yin, Yu, Tang, Yang, Xu, and Xiong]{li2025memosoperatingmemoryaugmentedgeneration}
Zhiyu Li, Shichao Song, Hanyu Wang, Simin Niu, Ding Chen, Jiawei Yang, Chenyang Xi, Huayi Lai, Jihao Zhao, Yezhaohui Wang, Junpeng Ren, Zehao Lin, Jiahao Huo, Tianyi Chen, Kai Chen, Kehang Li, Zhiqiang Yin, Qingchen Yu, Bo~Tang, Hongkang Yang, Zhi-Qin~John Xu, and Feiyu Xiong.
\newblock Memos: An operating system for memory-augmented generation (mag) in large language models, 2025{\natexlab{b}}.
\newblock URL \url{https://arxiv.org/abs/2505.22101}.

\bibitem[Liang et~al.(2026)Liang, Zeng, Zhao, and Zhao]{Liang_Zeng_Zhao_Zhao_2026}
Feng Liang, Weixin Zeng, Runhao Zhao, and Xiang Zhao.
\newblock Nestr: A neuro-symbolic abductive framework for temporal reasoning in large language models.
\newblock \emph{Proceedings of the AAAI Conference on Artificial Intelligence}, 40\penalty0 (38):\penalty0 31907--31915, Mar. 2026.
\newblock \doi{10.1609/aaai.v40i38.40460}.
\newblock URL \url{https://ojs.aaai.org/index.php/AAAI/article/view/40460}.

\bibitem[Luo et~al.(2025)Luo, Zhang, Yuan, Zhao, Yang, Gu, Wu, Chen, Qiao, Long, Tu, Luo, Ju, Xiao, Wang, Xiao, Liu, Yuan, Zhang, Jin, Zhang, Wu, Zhao, Tao, Yu, and Zhang]{luo2025largelanguagemodelagent}
Junyu Luo, Weizhi Zhang, Ye~Yuan, Yusheng Zhao, Junwei Yang, Yiyang Gu, Bohan Wu, Binqi Chen, Ziyue Qiao, Qingqing Long, Rongcheng Tu, Xiao Luo, Wei Ju, Zhiping Xiao, Yifan Wang, Meng Xiao, Chenwu Liu, Jingyang Yuan, Shichang Zhang, Yiqiao Jin, Fan Zhang, Xian Wu, Hanqing Zhao, Dacheng Tao, Philip~S. Yu, and Ming Zhang.
\newblock Large language model agent: A survey on methodology, applications and challenges, 2025.
\newblock URL \url{https://arxiv.org/abs/2503.21460}.

\bibitem[Maharana et~al.(2024)Maharana, Lee, Tulyakov, Bansal, Barbieri, and Fang]{maharana-etal-2024-evaluating}
Adyasha Maharana, Dong-Ho Lee, Sergey Tulyakov, Mohit Bansal, Francesco Barbieri, and Yuwei Fang.
\newblock Evaluating very long-term conversational memory of {LLM} agents.
\newblock In Lun-Wei Ku, Andre Martins, and Vivek Srikumar, editors, \emph{Proceedings of the 62nd Annual Meeting of the Association for Computational Linguistics (Volume 1: Long Papers)}, pages 13851--13870, Bangkok, Thailand, August 2024. Association for Computational Linguistics.
\newblock \doi{10.18653/v1/2024.acl-long.747}.
\newblock URL \url{https://aclanthology.org/2024.acl-long.747/}.

\bibitem[Matarazzo and Torlone(2025)]{matarazzo2025surveylargelanguagemodels}
Andrea Matarazzo and Riccardo Torlone.
\newblock A survey on large language models with some insights on their capabilities and limitations, 2025.
\newblock URL \url{https://arxiv.org/abs/2501.04040}.

\bibitem[Mei et~al.(2026)Mei, Chen, Yang, Hou, Li, and Byrne]{mei2026accordingmelongtermpersonalized}
Jingbiao Mei, Jinghong Chen, Guangyu Yang, Xinyu Hou, Margaret Li, and Bill Byrne.
\newblock According to me: Long-term personalized referential memory qa, 2026.
\newblock URL \url{https://arxiv.org/abs/2603.01990}.

\bibitem[Minaee et~al.(2025)Minaee, Mikolov, Nikzad, Chenaghlu, Socher, Amatriain, and Gao]{minaee2025largelanguagemodelssurvey}
Shervin Minaee, Tomas Mikolov, Narjes Nikzad, Meysam Chenaghlu, Richard Socher, Xavier Amatriain, and Jianfeng Gao.
\newblock Large language models: A survey, 2025.
\newblock URL \url{https://arxiv.org/abs/2402.06196}.

\bibitem[Packer et~al.(2024)Packer, Wooders, Lin, Fang, Patil, Stoica, and Gonzalez]{packer2024memgptllmsoperatingsystems}
Charles Packer, Sarah Wooders, Kevin Lin, Vivian Fang, Shishir~G. Patil, Ion Stoica, and Joseph~E. Gonzalez.
\newblock Memgpt: Towards llms as operating systems, 2024.
\newblock URL \url{https://arxiv.org/abs/2310.08560}.

\bibitem[Qian et~al.(2024)Qian, Zhang, Zhao, Zhou, Sui, Zhang, and Song]{qian-etal-2024-timer4}
Xinying Qian, Ying Zhang, Yu~Zhao, Baohang Zhou, Xuhui Sui, Li~Zhang, and Kehui Song.
\newblock {T}ime{R}$^4$ : Time-aware retrieval-augmented large language models for temporal knowledge graph question answering.
\newblock In Yaser Al-Onaizan, Mohit Bansal, and Yun-Nung Chen, editors, \emph{Proceedings of the 2024 Conference on Empirical Methods in Natural Language Processing}, pages 6942--6952, Miami, Florida, USA, November 2024. Association for Computational Linguistics.
\newblock \doi{10.18653/v1/2024.emnlp-main.394}.
\newblock URL \url{https://aclanthology.org/2024.emnlp-main.394/}.

\bibitem[Rasmussen et~al.(2025)Rasmussen, Paliychuk, Beauvais, Ryan, and Chalef]{rasmussen2025zeptemporalknowledgegraph}
Preston Rasmussen, Pavlo Paliychuk, Travis Beauvais, Jack Ryan, and Daniel Chalef.
\newblock Zep: A temporal knowledge graph architecture for agent memory, 2025.
\newblock URL \url{https://arxiv.org/abs/2501.13956}.

\bibitem[Robertson and Zaragoza(2009)]{robertson2009probabilistic}
Stephen Robertson and Hugo Zaragoza.
\newblock The probabilistic relevance framework: {BM25} and beyond.
\newblock \emph{Foundations and Trends in Information Retrieval}, 3\penalty0 (4):\penalty0 333--389, 2009.
\newblock \doi{10.1561/1500000019}.

\bibitem[Shao et~al.(2024)Shao, Wang, Zhu, Xu, Song, Bi, Zhang, Zhang, Li, Wu, and Guo]{shao2024deepseekmathpushinglimitsmathematical}
Zhihong Shao, Peiyi Wang, Qihao Zhu, Runxin Xu, Junxiao Song, Xiao Bi, Haowei Zhang, Mingchuan Zhang, Y.~K. Li, Y.~Wu, and Daya Guo.
\newblock Deepseekmath: Pushing the limits of mathematical reasoning in open language models, 2024.
\newblock URL \url{https://arxiv.org/abs/2402.03300}.

\bibitem[Sheng et~al.(2025)Sheng, Zhang, Ye, Wu, Zhang, Zhang, Peng, Lin, and Wu]{Sheng_2025}
Guangming Sheng, Chi Zhang, Zilingfeng Ye, Xibin Wu, Wang Zhang, Ru~Zhang, Yanghua Peng, Haibin Lin, and Chuan Wu.
\newblock {HybridFlow}: A flexible and efficient {RLHF} framework.
\newblock In \emph{Proceedings of the Twentieth European Conference on Computer Systems}, EuroSys '25, pages 1279--1297. ACM, March 2025.
\newblock \doi{10.1145/3689031.3696075}.
\newblock URL \url{http://dx.doi.org/10.1145/3689031.3696075}.

\bibitem[Song et~al.(2025)Song, Liang, Sun, Zhang, Wang, and Xu]{10.1145/3726302.3730173}
Xintong Song, Bin Liang, Yang Sun, Chenhua Zhang, Bingbing Wang, and Ruifeng Xu.
\newblock Bridging time gaps: Temporal logic relations for enhancing temporal reasoning in large language models.
\newblock In \emph{Proceedings of the 48th International ACM SIGIR Conference on Research and Development in Information Retrieval}, SIGIR '25, page 3040–3044, New York, NY, USA, 2025. Association for Computing Machinery.
\newblock ISBN 9798400715921.
\newblock \doi{10.1145/3726302.3730173}.
\newblock URL \url{https://doi.org/10.1145/3726302.3730173}.

\bibitem[Su et~al.(2026)Su, Guo, Hou, Bai, Li, Zhang, Yin, Lin, Jin, Guo, and Cheng]{su2026dialoguetimetemporalsemantic}
Miao Su, Yucan Guo, Zhongni Hou, Long Bai, Zixuan Li, Yufei Zhang, Guojun Yin, Wei Lin, Xiaolong Jin, Jiafeng Guo, and Xueqi Cheng.
\newblock Beyond dialogue time: Temporal semantic memory for personalized llm agents, 2026.
\newblock URL \url{https://arxiv.org/abs/2601.07468}.

\bibitem[Wang et~al.(2025)Wang, Takanobu, Liang, Mao, Hu, McAuley, and Wu]{wang2025memalphalearningmemoryconstruction}
Yu~Wang, Ryuichi Takanobu, Zhiqi Liang, Yuzhen Mao, Yuanzhe Hu, Julian McAuley, and Xiaojian Wu.
\newblock {Mem-$\alpha$: Learning Memory Construction via Reinforcement Learning}, 2025.
\newblock URL \url{https://arxiv.org/abs/2509.25911}.

\bibitem[Wu et~al.(2024)Wu, Liu, He, Liu, Zhang, Wang, and Wang]{10.1145/3627673.3679800}
Feifan Wu, Lingyuan Liu, Wentao He, Ziqi Liu, Zhiqiang Zhang, Haofen Wang, and Meng Wang.
\newblock Time-sensitive retrieval-augmented generation for question answering.
\newblock In \emph{Proceedings of the 33rd ACM International Conference on Information and Knowledge Management}, CIKM '24, page 2544–2553, New York, NY, USA, 2024. Association for Computing Machinery.
\newblock ISBN 9798400704369.
\newblock \doi{10.1145/3627673.3679800}.
\newblock URL \url{https://doi.org/10.1145/3627673.3679800}.

\bibitem[Wu and Hooi(2025)]{wu2025chainoftimeline}
Jiaying Wu and Bryan Hooi.
\newblock Chain-of-timeline: Enhancing {LLM} zero-shot temporal reasoning with {SQL}-style timeline formalization.
\newblock In \emph{Workshop on Reasoning and Planning for Large Language Models}, 2025.
\newblock URL \url{https://openreview.net/forum?id=ptHtrHwMpw}.

\bibitem[Xu et~al.(2026)Xu, Liang, Mei, Gao, Tan, and Zhang]{xu2026amem}
Wujiang Xu, Zujie Liang, Kai Mei, Hang Gao, Juntao Tan, and Yongfeng Zhang.
\newblock A-mem: Agentic memory for {LLM} agents.
\newblock In \emph{The Thirty-ninth Annual Conference on Neural Information Processing Systems}, 2026.
\newblock URL \url{https://openreview.net/forum?id=FiM0M8gcct}.

\bibitem[Yan et~al.(2026)Yan, Yang, Huang, Nie, Ding, Li, Ma, Bi, Kersting, Pan, Schütze, Tresp, and Ma]{yan2026memoryr1enhancinglargelanguage}
Sikuan Yan, Xiufeng Yang, Zuchao Huang, Ercong Nie, Zifeng Ding, Zonggen Li, Xiaowen Ma, Jinhe Bi, Kristian Kersting, Jeff~Z. Pan, Hinrich Schütze, Volker Tresp, and Yunpu Ma.
\newblock Memory-r1: Enhancing large language model agents to manage and utilize memories via reinforcement learning, 2026.
\newblock URL \url{https://arxiv.org/abs/2508.19828}.

\bibitem[Yang et~al.(2025)Yang, Li, Yang, Zhang, Hui, Zheng, Yu, Gao, Huang, Lv, Zheng, Liu, Zhou, Huang, Hu, Ge, Wei, Lin, Tang, Yang, Tu, Zhang, Yang, Yang, Zhou, Zhou, Lin, Dang, Bao, Yang, Yu, Deng, Li, Xue, Li, Zhang, Wang, Zhu, Men, Gao, Liu, Luo, Li, Tang, Yin, Ren, Wang, Zhang, Ren, Fan, Su, Zhang, Zhang, Wan, Liu, Wang, Cui, Zhang, Zhou, and Qiu]{yang2025qwen3technicalreport}
An~Yang, Anfeng Li, Baosong Yang, Beichen Zhang, Binyuan Hui, Bo~Zheng, Bowen Yu, Chang Gao, Chengen Huang, Chenxu Lv, Chujie Zheng, Dayiheng Liu, Fan Zhou, Fei Huang, Feng Hu, Hao Ge, Haoran Wei, Huan Lin, Jialong Tang, Jian Yang, Jianhong Tu, Jianwei Zhang, Jianxin Yang, Jiaxi Yang, Jing Zhou, Jingren Zhou, Junyang Lin, Kai Dang, Keqin Bao, Kexin Yang, Le~Yu, Lianghao Deng, Mei Li, Mingfeng Xue, Mingze Li, Pei Zhang, Peng Wang, Qin Zhu, Rui Men, Ruize Gao, Shixuan Liu, Shuang Luo, Tianhao Li, Tianyi Tang, Wenbiao Yin, Xingzhang Ren, Xinyu Wang, Xinyu Zhang, Xuancheng Ren, Yang Fan, Yang Su, Yichang Zhang, Yinger Zhang, Yu~Wan, Yuqiong Liu, Zekun Wang, Zeyu Cui, Zhenru Zhang, Zhipeng Zhou, and Zihan Qiu.
\newblock Qwen3 technical report, 2025.
\newblock URL \url{https://arxiv.org/abs/2505.09388}.

\bibitem[Zhang et~al.(2025{\natexlab{a}})Zhang, Xue, Zhang, Wu, Luu, and Zhao]{zhang-etal-2025-mrag}
Siyue Zhang, Yuxiang Xue, Yiming Zhang, Xiaobao Wu, Anh~Tuan Luu, and Chen Zhao.
\newblock {MRAG}: A modular retrieval framework for time-sensitive question answering.
\newblock In Christos Christodoulopoulos, Tanmoy Chakraborty, Carolyn Rose, and Violet Peng, editors, \emph{Findings of the Association for Computational Linguistics: EMNLP 2025}, pages 3080--3118, Suzhou, China, November 2025{\natexlab{a}}. Association for Computational Linguistics.
\newblock ISBN 979-8-89176-335-7.
\newblock \doi{10.18653/v1/2025.findings-emnlp.167}.
\newblock URL \url{https://aclanthology.org/2025.findings-emnlp.167/}.

\bibitem[Zhang et~al.(2025{\natexlab{b}})Zhang, Rossi, Kveton, Shao, Yang, Zamani, Dernoncourt, Barrow, Yu, Kim, Zhang, Gu, Derr, Chen, Wu, Chen, Wang, Mitra, Lipka, Ahmed, and Wang]{zhang2025personalizationlargelanguagemodels}
Zhehao Zhang, Ryan~A. Rossi, Branislav Kveton, Yijia Shao, Diyi Yang, Hamed Zamani, Franck Dernoncourt, Joe Barrow, Tong Yu, Sungchul Kim, Ruiyi Zhang, Jiuxiang Gu, Tyler Derr, Hongjie Chen, Junda Wu, Xiang Chen, Zichao Wang, Subrata Mitra, Nedim Lipka, Nesreen Ahmed, and Yu~Wang.
\newblock Personalization of large language models: A survey, 2025{\natexlab{b}}.
\newblock URL \url{https://arxiv.org/abs/2411.00027}.

\bibitem[Zhong et~al.(2024)Zhong, Guo, Gao, Ye, and Wang]{Zhong_Guo_Gao_Ye_Wang_2024}
Wanjun Zhong, Lianghong Guo, Qiqi Gao, He~Ye, and Yanlin Wang.
\newblock Memorybank: Enhancing large language models with long-term memory.
\newblock \emph{Proceedings of the AAAI Conference on Artificial Intelligence}, 38\penalty0 (17):\penalty0 19724--19731, Mar. 2024.
\newblock \doi{10.1609/aaai.v38i17.29946}.
\newblock URL \url{https://ojs.aaai.org/index.php/AAAI/article/view/29946}.

\end{thebibliography}
